\documentclass[11pt]{article}
\usepackage{acl}
\usepackage{times}
\usepackage{latexsym}
\usepackage[T1]{fontenc}
\usepackage[utf8]{inputenc}
\usepackage{microtype}
\usepackage{multirow}
\usepackage[normalem]{ulem}
\useunder{\uline}{\ul}{}
\usepackage{makecell}
\usepackage{bm}
\usepackage{cite}
\usepackage{setspace}
\usepackage{algorithmic}
\usepackage{algorithm}
\usepackage{threeparttable}
\usepackage{subfigure}
\usepackage{amstext}
\usepackage{booktabs}
\usepackage{graphicx}
\newcommand{\para}[1]{\vspace{.05in}\noindent\textbf{#1}}
\usepackage{authblk}

\title{Explore More Guidance: A Task-aware Instruction Network  \\for Sign Language Translation Enhanced with Data Augmentation}

\makeatletter
\def\thanks#1{\protected@xdef\@thanks{\@thanks
        \protect\footnotetext{#1}}}
\makeatother

\newcommand*{\affaddr}[1]{#1} 
\newcommand*{\affmark}[1][*]{\textsuperscript{#1}}
\newcommand*{\email}[1]{\normalsize {#1}}

\author{\bf Yong Cao\affmark[1*] \thanks{* Equal Contribution.}, \bf Wei Li\affmark[2*], \bf Xianzhi Li\affmark[1*], \bf Min Chen\affmark[1\#] \thanks{\# Corresponding author: Min Chen.}, \bf Guangyong Chen\affmark[3] \\
        \bf Long Hu\affmark[1], \bf Zhengdao Li\affmark[4] and \bf Hwang Kai\affmark[4] \\
\affaddr{\affmark[1]Huazhong University of Science and Technology}, \affaddr{\affmark[2]Nanchang University}\\
\affaddr{\affmark[3]Zhejiang University, Zhejiang Lab}, \affaddr{\affmark[4]The Chinese University of Hong Kong, Shenzhen}\\
\email{\{yongcao\_epic, weili\_epic, xzli, minchen2012\}@hust.edu.cn, gychen@zhejianglab.com} \\
\email{hulong@hust.edu.cn, zhengdaoli@link.cuhk.edu.cn, hwangkai@cuhk.edu.cn} \\
}

\begin{document}
\maketitle
\begin{abstract}
Sign language recognition and translation first uses a recognition module to generate glosses from sign language videos and then employs a translation module to translate glosses into spoken sentences.
Most existing works focus on the recognition step, while paying less attention to sign language translation.
In this work, we propose a task-aware instruction network, namely TIN-SLT, for sign language translation, by introducing the instruction module and the learning-based feature fuse strategy into a Transformer network.
In this way, the pre-trained model's language ability can be well explored and utilized to further boost the translation performance.
Moreover, by exploring the representation space of sign language glosses and target spoken language, we propose a multi-level data augmentation scheme to adjust the data distribution of the training set.
We conduct extensive experiments on two challenging benchmark datasets, PHOENIX-2014-T and ASLG-PC12, on which our method outperforms former best solutions by 1.65 and 1.42 in terms of BLEU-4.
Our code is published at \url{https://github.com/yongcaoplus/TIN-SLT}.
\end{abstract}

\section{Introduction}\label{sec:Introduction}

\begin{figure}[ht]
\centering
\includegraphics[width=1\columnwidth]{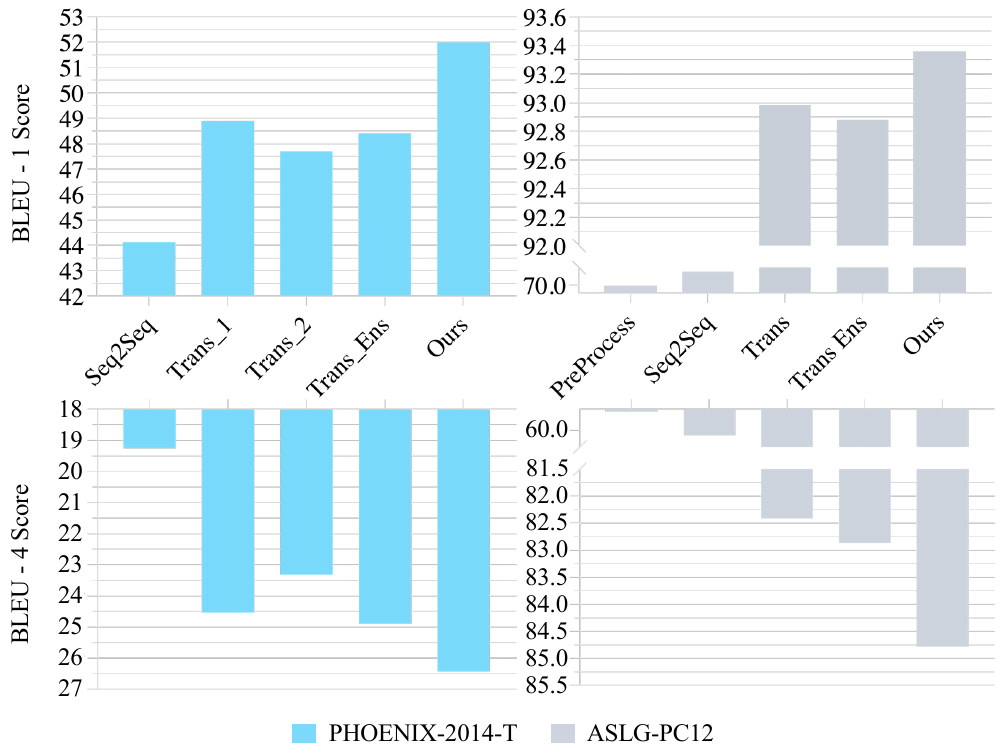}
\caption{Comparing the sign language translation performance on two challenging datasets, i.e., PHOENIX-2014-T (blue) and ASLG-PC12 (gray), in terms of BLEU-1 and BLEU-4 metrics. Clearly, our approach achieves the highest scores on both datasets compared with others. The experiments section contains more results and analysis.}
\label{fig:result_compare}
\vspace*{-3mm}
\end{figure}

Sign language recognition and translation aims to transform sign language videos into spoken languages, which builds a bridge for communication between deaf and normal people.
Considering the unique grammar of sign languages, current effective recognition and translation systems involve two steps: a tokenization module to generate glosses from sign language videos, and a translation module to translate the recognized glosses into spoken natural languages.
Previous works~\citep{li2020word, sincan2020autsl, sharma2021asl, kumar20203d, camgoz2020sign} have proposed various solutions to address the first step, but paid less attention to the translation system.
Hence, this paper aims to solve the problem of sign language translation (SLT) with the goal of translating multiple recognized independent glosses into a complete sentence.

To do so, most existing works \citep{ko2019neural, stoll2018sign} directly apply advanced techniques, e.g., Seq2Seq model \citep{sutskever2014sequence} or Transformer \citep{vaswani2017attention}, from neural machine translation to SLT.
However, different from the lingual translation task in neural machine translation, SLT poses several unique challenges.
First, it is hard to collect and annotate a large amount of sign language corpus.
It is still an open question that how to explore more guidance and external information for SLT task by incorporating the pre-trained language models based on masses of unlabeled corpus.
Second, since sign languages are developed independently from spoken languages with quite different linguistic features, the discrepancy of representation space between glosses and spoken sentences is significant, thus increasing the translation difficulty.

To address the above issues, we propose a novel task-aware instruction network, called TIN-SLT for sign language translation, further enhanced with a multi-level data augmentation scheme.
Our TIN-SLT is capable of encoding pre-trained language model's ability into the translation model and also decreasing the discrepancy between the representation space of glosses and texts.

\textbf{To begin with}, we leverage the extracted hidden features from the pre-trained model as extra information to guide the sign language translation. Besides, we apply an instruction module to transform general token features into task-aware features. In this way, we can fully utilize the language skills originating from the external world, thus reducing the demand for sign language training data.

\textbf{Next}, to better inject the information from pre-trained model into the SLT model, we design a learning-based feature fusion strategy, which has been analyzed and validated to be effective compared with existing commonly-used fusion ways.

\textbf{Finally}, considering the large difference between the sign language glosses and texts in terms of the representation space, we propose a multi-level data augmentation scheme to enrich the coverage and variety of existing datasets.

In summary, our contributions are threefold:
(i) a novel TIN-SLT network to explore more guidance of pre-trained models,
(ii) a learning-based feature fusion strategy,
and (iii) a multi-level data augmentation scheme.
Extensive experiments on challenging benchmark datasets validate the superiority of our TIN-SLT over state-of-the-art approaches; see Figure~\ref{fig:result_compare} for example results.
\section{Related Works}
\label{sec:RelatedWorks}

\para{Methods for sign language recognition.} \
SLR task mainly focuses on the extraction of extended spatial and temporal multi-cue features \citep{zhou2020spatial, Koller_2017_CVPR}. Most existing works \citep{yin2016iterative, qiu2017learning, wei2019deep, cui2019deep} study the strong representation of sign language videos such as multi-semantic \citep{cui2019deep} and multi-modality \citep{koller2019weakly} analysis.
Although extracting representative features from sign language videos is fully explored, how to effectively conduct the subsequent translation by considering the unique linguistic features of sign language is often ignored in these SLR works.

\para{Methods for sign language translation.} \
Early approaches for SLT rely on seq2seq model and attention mechanism \citep{arvanitis2019translation}, while facing the limitation of long-term dependencies.
Later, motivated by the ability of the Transformer \citep{vaswani2017attention}, many researchers utilize it to effectively improve SLT performance.
For example, the work in \citet{camgoz2020sign} tried to use Transformer for both recognition and translation, and promote the joint optimization of sign language recognition and translation.
The subsequent work \citep{yin2020better} proposed the STMC-Transformer network which first uses STMC networks \citep{zhou2020spatial} to achieve better results for SLR, and then exploits Transformer for translation to obtain better SLT performance.

\begin{figure}[t]
	\centering
	\includegraphics[width=1.0\columnwidth]{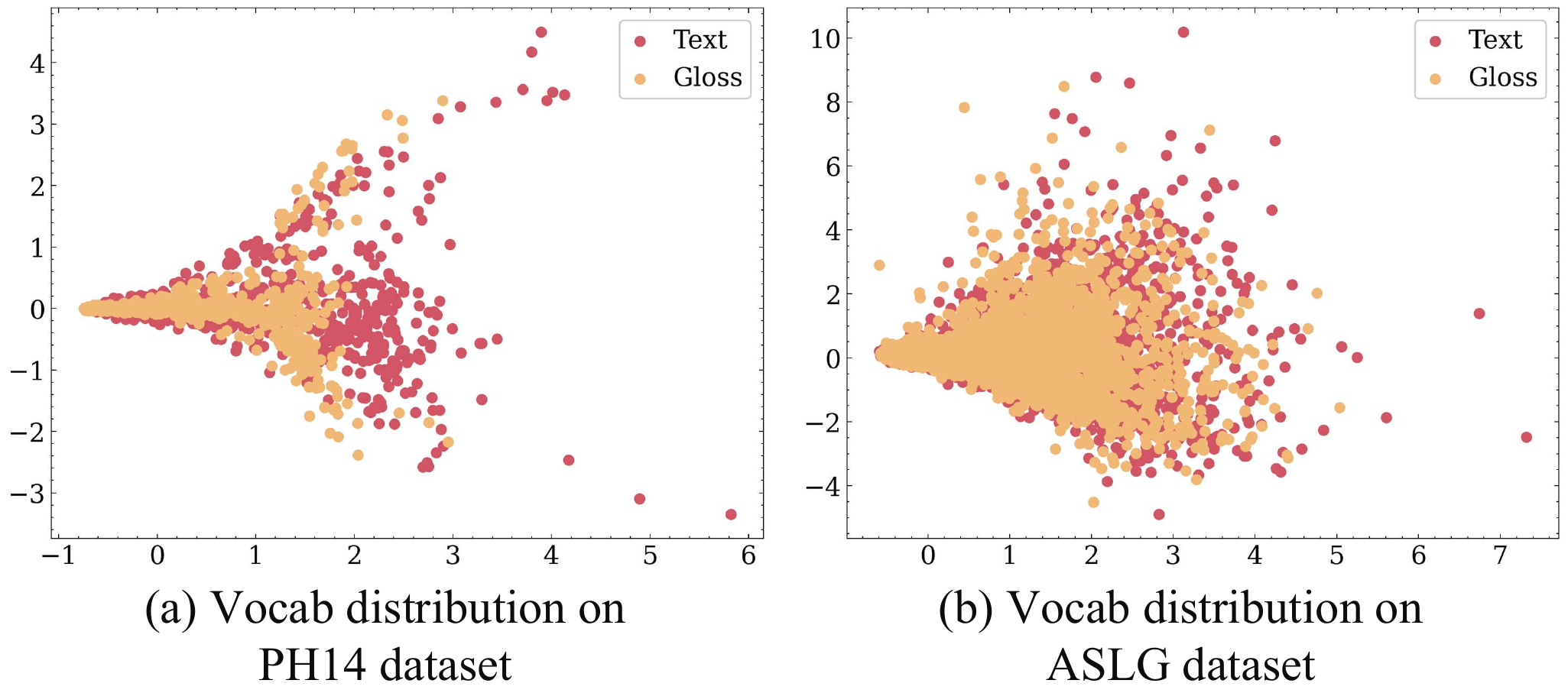}
	\caption{Comparing the sample distribution between the input sign glosses (yellow dots) and the output translated texts (red dots) on two datasets.}
	\label{fig:dataset_distribution}
\end{figure}

\para{General neural machine translation.} \
Broadly speaking, sign language translation belongs to the field of neural machine translation, with the goal of carrying out automated text translation.
Earlier approaches deployed recurrent network \citep{bahdanau2014neural}, convolutional network \citep{gehring2017convolutional}, or Transformer \citep{vaswani2017attention} as encoder-decoder module.
Among them, Transformer has achieved state-of-the-art results, but the translation performance still needs to be improved due to the limited training corpus. In addition, there are some explorations in bringing the pre-trained models into neural machine translation \citep{imamura2019recycling, shavarani2021better, zhu2020incorporating}.

\section{Challenges}\label{sec:Challenges}
The goal of this work is to translate the recognized multiple independent glosses (network input) into a complete spoken sentence (expected output).
Compared with general neural machine translation tasks, SLT faces two main challenges:

\begin{figure*}[t]
	\centering
	\includegraphics[width=2\columnwidth]{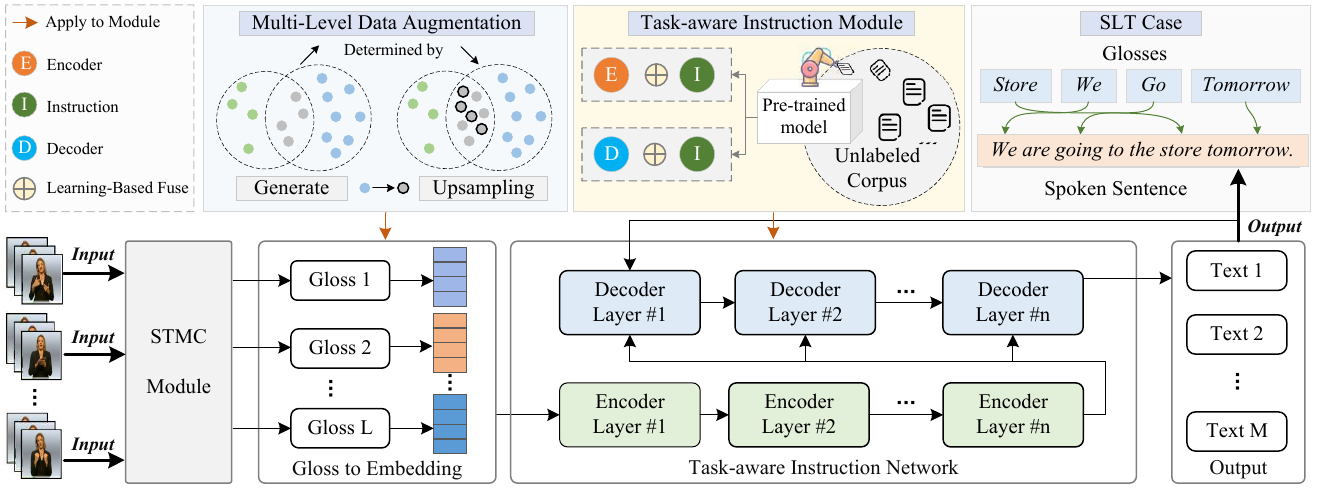}
	\caption{Network architecture of TIN-SLT. As shown in the bottom row, we first employ STMC model \citep{zhou2020spatial} to recognize sign language videos to independent glosses. Next, we design a multi-level data augmentation scheme to enrich existing data pool for better feature embedding from glosses. Then, we design a task-aware instruction network with a novel instruction module to translate glosses into a complete spoken sentence.}

	\label{fig:architecture}
\end{figure*}

\para{Limited annotated corpus}: \
Compared with natural languages, the data resources of sign languages are scarce~\citep{bragg2019sign}.
As a result, the SLT models trained on limited data often suffer from the overfitting problem with poor generalization~\citep{2021Data, 2021Including}.

\para{Discrepancy between glosses (input) and texts (output)}: \
Figure \ref{fig:dataset_distribution} shows the representation space of sign glosses (yellow dots) and translated texts (red dots) using Word2Vec \citep{mikolov2013distributed} on two different datasets.
We can observe that the representation space of sign glosses is clearly smaller than that of the target spoken language, thus increasing the difficulty of network learning.

\section{Our Approach}\label{sec:HC}
To address the above challenges, we propose TIN-SLT by effectively introducing the pre-trained model into SLT task and further designing a multi-level data augmentation scheme.
Figure~\ref{fig:architecture} depicts the detailed network architecture.
In the following subsections, we will firstly introduce the network architecture of TIN-SLT, followed by our solutions to address the above two challenges.

\subsection{Network Architecture of TIN-SLT}

Given a sign language video $\mathcal{V}=\{V_1,\dots,V_T\}$ with $T$ frames, like existing approaches, we also adopt a two-step pipeline by first (i) recognizing  $\mathcal{V}$ into a sequence $\mathcal{G}=\{g_1,\dots,g_L\}$ with $L$ independent glosses and then (ii) translating $\mathcal{G}$ into a complete spoken sentence $\mathcal{S}=\{w_1,\dots,w_M\}$ with $M$ words, but we pay more attention to solve step (ii).
Hence, for step (i), as shown in the bottom-left part of Figure~\ref{fig:architecture}, we empirically use the spatial-temporal multi-cue (STMC) network \citep{zhou2020spatial}, which consists of a spatial multi-cue module and a temporal multi-cue module.
For more technical details of STMC, please refer to~\citep{zhou2020spatial}.
Below, we shall mainly elaborate on the details of addressing step (ii).

After obtaining the sequence $\mathcal{G}$ of sign glosses, considering that the representation space of glosses is much smaller than that of texts (see Figure~\ref{fig:dataset_distribution}), we thus design a multi-level data augmentation scheme to expand the gloss representation space; see the top-left part of Figure~\ref{fig:architecture} as an illustration and we shall present its details in Section~\ref{subsec:augmentation}.

Next, as shown in the bottom-middle part of Figure~\ref{fig:architecture}, the key of our design is a task-aware instruction network, where we adopt Transformer as the network backbone consisting of several encoder and decoder layers, whose objective is to learn the conditional probabilities $p(\mathcal{S}|\mathcal{G})$.
Since SLT is an extremely low-data-resource task as we have discussed in Section~\ref{sec:Challenges}, we thus focus on exploring more task-aware guidance by learning external world knowledge, which is dynamically incorporated into the Transformer backbone via our designed task-aware instruction module.
We shall present its details in Section~\ref{subsec:TIM}.

Lastly, the outputs of last decoder are passed through a non-linear point-wise feed forward layer and we can obtain the predicted sentence $\mathcal{S}$ by a linear transform and softmax layer.

\subsection{Task-aware Instruction Module}
\label{subsec:TIM}
As is shown in Figure~\ref{fig:architecture}, our task-aware instruction network is composed of a series of encoder and decoder layers.
To handle the limited training data, we propose to leverage the learned external knowledge from natural language datasets to guide the learning of sign languages.
More specifically, we design a task-aware instruction module to dynamically inject external knowledge from pre-trained models into our encoder and decoder.
Below, we shall present the details.

\begin{figure}[t]
\centering
\includegraphics[width=1\columnwidth]{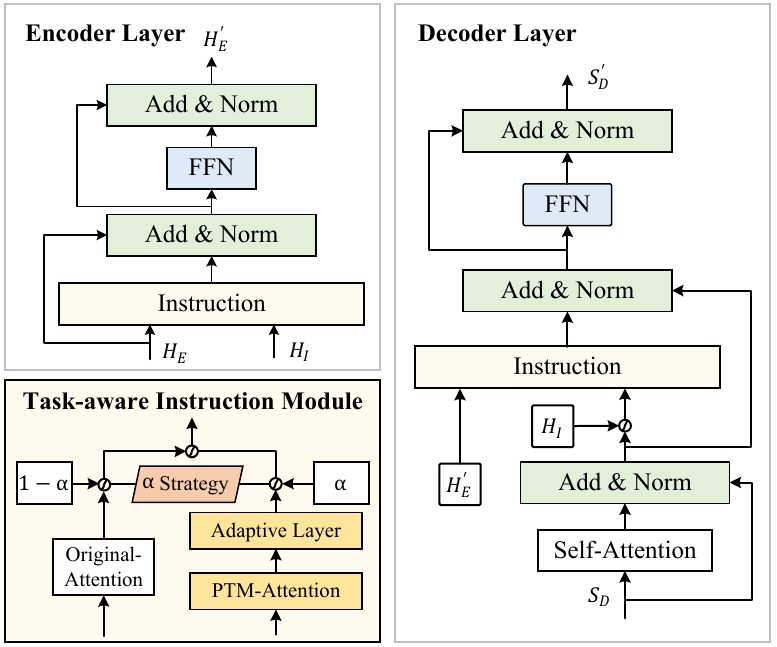}
\caption{Details of Encoder layer, Decoder layer, and and Instruction Module.}
\label{fig:instructed-attention}
\end{figure}

\para{Encoder.} \
Given the recognized glosses,let $H_I$ denotes the instruction features encoded by the pre-trained model (PTM), $H_E$ and $H_E^{\prime}$ denotes the input and output of encoder which is randomly initialized.
As shown in Figure~\ref{fig:instructed-attention}, $H_I$ and $H_E$ are fed into the task-aware instruction module for feature fusing.
Then, the output of the instruction module is fed into residual connection (Add\&Norm) and feed forward network (FFN).

The light yellow box of Figure~\ref{fig:instructed-attention} shows the detailed design of task-aware instruction module.
Specifically, we feed $H_E$ into a self-attention module to learn the contextual relationship between the features of glosses, while $H_I$ is fed into a PTM-attention, which is the same architecture as self-attention. Different from existing work which employ PTM in general neural network~\citep{zhu2020incorporating}, we insert an adaptive layer to fine-tune PTM-attention output for SLT task, to transform general gloss features into task-aware features.

%
\begin{equation}
h_i = \sigma ( Attn_I(h_t, H_I, H_I))
\end{equation}
where $\sigma()$ denotes the adaptive layer (we set it as fully connection layers here), and $h_t$ denotes the gloss features at time step $t$. Then, the output of two modules are combined via $\alpha$ strategy.
The whole process is formulated as follows:
\begin{equation}\label{eq:fuse}
	\hat{h}_t = (1 - \alpha) Attn_E(h_t, H_E, H_E) + \alpha h_i
\end{equation}
where $Attn_E$ and $Attn_I$ are two attention layers with different parameters, which follow \citep{vaswani2017attention}.
The way of setting an optimal $\alpha$ will be introduced later.

\para{Decoder.} \
Let $S_D$ and $S^{\prime}_D$ denotes the input and output of decoder, $s_t$ denote the hidden state at time step $t$, and $s_0$ denotes the beginning token of a sentence, i.e., $<bos>$.
The hidden states are passed to a masked self-attention ensuring that each token may only use its predecessors as follows:

\begin{equation}
	\tilde{s}_t = Attn_D(s_t, s_{1:t}, s_{1:t})
\end{equation}

Representations $H_E^{\prime}$ and $H_I$ extracted from encoder and PTM are fed into the decoder-attention and PTM-attention module, respectively, as shown in the right part of Figure~\ref{fig:instructed-attention}. Similar to Encoder, we formulate this decoding output as:

\begin{equation}\label{eq:fuse_dec}
	\hat s_t =  (1-\alpha) Attn_D(\tilde{s}_t, H_E^{\prime}, H_E^{\prime}) + \alpha h_i
\end{equation}

where $Attn_D$ represent decoder-attention, and $\hat s_t$ is the output of decoder instruction module.

\para{Learning-based feature fusion.} \
As shown in Eq. (\ref{eq:fuse}), representations extracted from both PTM- and self- attention are fused via a parameter $\alpha$.
How to set a reasonable and optimal $\alpha$ will directly affects the learning performance, which is a problem worthy of exploration.
Instead of manually setting a constant $\alpha$, we propose a learning-based strategy to encourage the network to learn the optimal $\alpha$ by itself for better feature fusion.

Specifically, learning-based strategy means that we adopt the back-propagation learning algorithm to update $\alpha$ during the network training process:

\begin{equation}\label{equ:alpha_learn}
	\alpha_{t+1} = \Gamma (\alpha_t, g_t)
	\label{alpha}
\end{equation}
where $g_t$ indicates the gradient and $\Gamma(\cdot)$ represents the optimization algorithm.
Though the idea of self-learning is straightforward, we shall show in the experiment section that it is quite effective compared with many other strategies.

\subsection{Multi-level Data Augmentation}
\label{subsec:augmentation}
To decrease the discrepancy between glosses (input) and texts (output), we propose a multi-level data augmentation scheme.
Our key idea is that, besides existing gloss-text pairs, we use upsampling as our data augmentation algorithm and generate text-text pairs as extended samples to introduce texts information into glosses, thus enlarging the feature distribution space of glosses.

Actually, there is a trade-off between augmentation and overfitting, which means the upsampling ratio $\Phi_{upsamp}$ should be determined by the degree of gloss-text difference.
%
We here propose four factors $\phi = [\phi_v, \phi_r, \phi_s, \phi_d]$ to calculate the difference in terms of token, sentence and dataset level, and set weighted $\phi$ as $\Phi_{upsamp}$.

\para{Token level.} \
Vocabulary Different Ratio (VDR, $\phi_v$) is used to measure the difference of gloss vocabulary space and text's, as calculated by Eq. (\ref{equ:vdr}).
\begin{equation}\label{equ:vdr}
	\phi_v = 1 - \frac{ | W_\mathcal{G} |}{|W_\mathcal{G} \cup W_\mathcal{S}|}
\end{equation}
where $W_\mathcal{G}$ and $W_\mathcal{S}$ represent gloss and text vocabularies, and $|\cdot|$ denotes the size of set.

We present Rare Vocabulary Ratio (RVR, $\phi_r$) to calculate the ratio of the rare words:
\begin{equation}\label{equ:rvr}
	\phi_r = 1 - \frac{ \sum_{\mathcal{G} \in W_{\mathcal{G}}}\#(Counter(\mathcal{G}) < \tau_{r})}{|W_\mathcal{G} \cup W_\mathcal{S}|}
\end{equation}
where $\#(\cdot)$ is 1 if the value is true, else 0, $Counter(\mathcal{G})$ is to calculate the gloss vocabulary frequency, and $\tau_{r}$ means the empirical thresh frequency determined by the vocabulary frequency, which is empirically set to be 2.

\para{Sentence level.} \
We propose Sentence Cover Ratio (SCR, $\phi_s$) to compute the gloss-text pair similarity and covered ratio, calculated as:
\begin{equation}
	r_i = \frac{|\mathcal{G}_i \cap \mathcal{S}_i|}{|\mathcal{S}_i|} , \quad
	\phi_s = 1 - \frac{1}{N} \sum_{i,r_i > \tau_{c}} r_i
\end{equation}
where $r_i$ denotes the covered ratio of gloss-text pair $\mathcal{G}_i$ and  $\mathcal{S}_i$, while $\tau_{c}$ means the empirical thresh (set $\tau_{c}=0.5$). We labeled gloss-text pairs which satisfy $r_i > \tau_{c}$ as candidates $\mathcal{C}$.

\para{Dataset level.} \
We use Dataset Length-difference Ratio (DLR, $\phi_d$) to calculate the length of sentence distance, calculated as:
\begin{equation}\label{equ:rvr}
	\phi_d = 1 - \frac{\sum_i |\mathcal{G}_i|}{\sum_i |\mathcal{S}_i|}
\end{equation}
Then we can get the upsampling ratio by:
\begin{equation}\label{equ:rvr}
	\Phi_{upsamp} = \theta * \phi
\end{equation}
where the weight matrix $\theta$ is empirically set as $ [0.1, 0.1, 0.6, 0.2]$, corresponding to the weight of $[\phi_v, \phi_r, \phi_s, \phi_d]$, as we suppose the sentence level matters the most and the weight of token level is the same as dataset level. Lastly, we obtain the upsampling ratio and use upsampling strategy among all candidates $\mathcal{C}$ to enrich the dataset.

\section{Experiments}\label{sec:Experiment}

\linespread{1.8}
\begin{table*}[t]
	\begin{spacing}{1.35}
		\resizebox{\textwidth}{!}{
			\begin{tabular}{l|llllll|llllll}
				\Xhline{1pt}
				\multicolumn{1}{c|}{}  & \multicolumn{6}{c|}{Dev Set}     & \multicolumn{6}{c}{Test Set}   \\ \cline{2-13}
				\multicolumn{1}{c|}{\multirow{-2}{*}{\textbf{Model}}}    & \multicolumn{1}{c}{BLEU-1}    & \multicolumn{1}{c}{BLEU-2}    & \multicolumn{1}{c}{BLEU-3}    & \multicolumn{1}{c}{BLEU-4}
				& \multicolumn{1}{c}{ROUGE-L}                           & METEOR                        & \multicolumn{1}{c}{BLEU-1}    & \multicolumn{1}{c}{BLEU-2}    & \multicolumn{1}{c}{BLEU-3}
				& \multicolumn{1}{c}{BLEU-4}                            & \multicolumn{1}{c}{ROUGE-L}   & METEOR                        \\  \hline \hline
				&\multicolumn{12}{c}{\textbf{PHOENIX-2014-T Dataset Evaluation}}                         \\
				Raw Data  (Yin and Read 2020)              &  13.01  &  6.23   &   3.03  &  1.71   &  24.23  &  13.69  &  11.88  &  5.05   &  2.41   &  1.36   &  22.81  &  12.12    \\
				Seq2seq  (Camgoz et al. 2018)            &  44.40  &  31.93  &  24.61  &  20.16  &  46.02  &  -      &  44.13  &  31.47  &  23.89  &  19.26  &  45.45  &  -        \\
				Transformer (Camgoz et al. 2020)          &  50.69  &  38.16  &  30.53  &  25.35  &  -      &  -      &  48.90  &  36.88  &  29.45  &  24.54  &  -      &  -        \\
				Transformer (Yin and Read 2020)          &  49.05  &  36.20  &  28.53  &  23.52  &  47.36  &  46.09  &  47.69  &  35.52  &  28.17  &  23.32  &  46.58  &  44.85    \\
				Transformer Ens. (Yin and Read 2020)      &  48.85  &  36.62  &  29.23  &  24.38  &  49.01  &  46.96  &  48.40  &  36.90  &  29.70  &  24.90  &  48.51  &  46.24    \\
				DataAug (Moryossef et al. 2021b)           &  -      &  -      &  -      &  -      &  -      &  -      &  -      &  -      &  -      &  23.35  &  -      &  -        \\  \hline
                \textbf{TIN-SLT(Ours)}   &  \textbf{52.35}   & \textbf{39.03}    & \textbf{30.83}    & \textbf{25.38}    & 48.82    & \textbf{48.40}
				&  \textbf{52.77}   & \textbf{40.08}    & \textbf{32.09}    & \textbf{26.55}    & \textbf{49.43}    & \textbf{49.36}     \\ \hline
				&\multicolumn{12}{c}{\textbf{ASLG-PC12 Dataset Evaluation}}                               \\
				Raw data (Yin and Read 2020)              &  54.60  &  39.67  &  28.92  &  21.16  &  76.11  &  61.25  &  54.19  &  39.26  &  28.44  &  20.63  &  75.59  &  61.65    \\
				Preprocessed data (Yin and Read 2020)     &  69.25  &  56.83  &  46.94  &  38.74  &  83.80  &  78.75  &  68.82  &  56.36  &  46.53  &  38.37  &  83.28  &  79.06    \\
				Seq2seq  (Arvanitis et al. 2019)    &  -      &  -      &  -      &  -      &  -      &  -      &  86.70  &  79.50  &  73.20  &  65.90  &  -      &  -        \\
				Transformer (Yin and Read 2020)           &  \textbf{92.98}  &  89.09  &  83.55  &  \textbf{85.63}   &  82.41  &  95.93   &  92.98  &  89.09  &  85.63  &  82.41
				&  95.87   &  96.46    \\
				Transformer Ens.(Yin and Read 2020)       &  92.67  &	88.72  &  85.22  & 	81.93  &  \textbf{96.18}  &  \textbf{95.95}  &  92.88  & 	89.22  &  85.95  &  82.87  &  \textbf{96.22}  &  \textbf{96.60}     \\  \hline
                \textbf{TIN-SLT (Ours)}  & 92.75   & \textbf{88.91}   & \textbf{85.51}   & 82.33   & 95.17    & 95.21   &  \textbf{93.35}  &  \textbf{90.03}  &  \textbf{87.07}  &  \textbf{84.29}
				& 95.39   & 95.92   \\ \Xhline{1pt}
			\end{tabular}
		}
	\caption{\label{G2T_result} Comparing the translation performance of TIN-SLT against state-of-the-art techniques on PHOENIX-2014-T and ASLG-PC12 datasets. Clearly, our TIN-SLT achieves the best performance on most metrics. }	
	\end{spacing}
\end{table*}

\subsection{Implementation Details}\label{sec:dataset}

\para{Datasets.} \
We conduct our experiments on two popular benchmark datasets of different languages and scales, including PHOENIX-2014-T \citep{camgoz2018neural} dataset and ASLG-PC12 \citep{othman2012english} dataset.

Specifically, PHOENIX-2014-T, i.e., PH14, is an open-source German sign language dataset, recorded from broadcast news about the weather.
This dataset contains parallel sign language videos from 9 different signers, gloss annotations with a vocabulary of 1066 different signs, and their translations with a vocabulary of 2887 different words.

ASLG-PC12, i.e., ASLG, is a parallel corpus of English written texts and American Sign Language (ASL) glosses, which is constructed based on rule-based approach.
It contains more than one hundred million pairs of sentences between English sentences and ASL glosses.

\para{Evaluation metrics.} \
To fairly evaluate the effectiveness of our TIN-SLT, we follow \citep{yin2020better} to use the commonly-used BLEU-$N$ ($N$-grams ranges from 1 to 4) \citep{papineni2002bleu}, ROUGE-L \citep{lin2004rouge} and METEOR \citep{banerjee2005meteor} as the evaluation metrics.

\para{Experimental setup.} \
The experiments are conducted on Ubuntu 18.04 system with two NVIDIA V100 GPUs.
Our Transformers are built using 2048 hidden units and 8 heads in each layer.
Besides, we adopt Adam \citep{kingma2014adam} as optimization algorithm with $\beta_1=0.9$, $\beta_2=0.998$ and use inverse sqrt learning rate scheduler with a weight decay of $10^{-3}$. Please refer to Appendix for more hyper-parameter settings.

\subsection{Comparison with Others}
To compare our TIN-SLT against state-of-the-art approaches on sign language translation task,
we conducted two groups of experiments, Gloss2Text (G2T) and Sign2Gloss2Text (S2G2T).

\para{Evaluation on G2T.} \
G2T is a text-to-text translation task, whose objective is to translate ground-truth sign glosses to spoken language sentences.
In specific, for PH14 dataset, we should output German spoken language sentences; while for ASLG dataset, we should output English sentences.
Table \ref{G2T_result} summarizes the comparison results.
Clearly, our TIN-SLT achieves the highest values on most evaluation metrics with a significant margin.
Particularly, the superiority of our method on PH14 dataset is more obvious, where almost all the evaluation values are the highest.
Thanks to our multi-level data augmentation scheme, the integrity of translated sentences has been improved, which is reflected in the significant improvement of BLEU-$N$ metric.
In addition, the strong guidance from external knowledge also encourages our network to generate translated sentences in correct grammar, consistent tense and appropriate word order.
For the lower ROUGE-L metric, we think that although the instruction module obviously help improve the accuracy and fluency of translation results, it leads to a slight decrease of continuous texts' recall rate in this task.

\begin{figure*}[t]
	\centering
	
	\subfigure[Comparing various $\alpha$ strategies on PH14 dataset]{%
		\includegraphics[width=0.46\columnwidth]{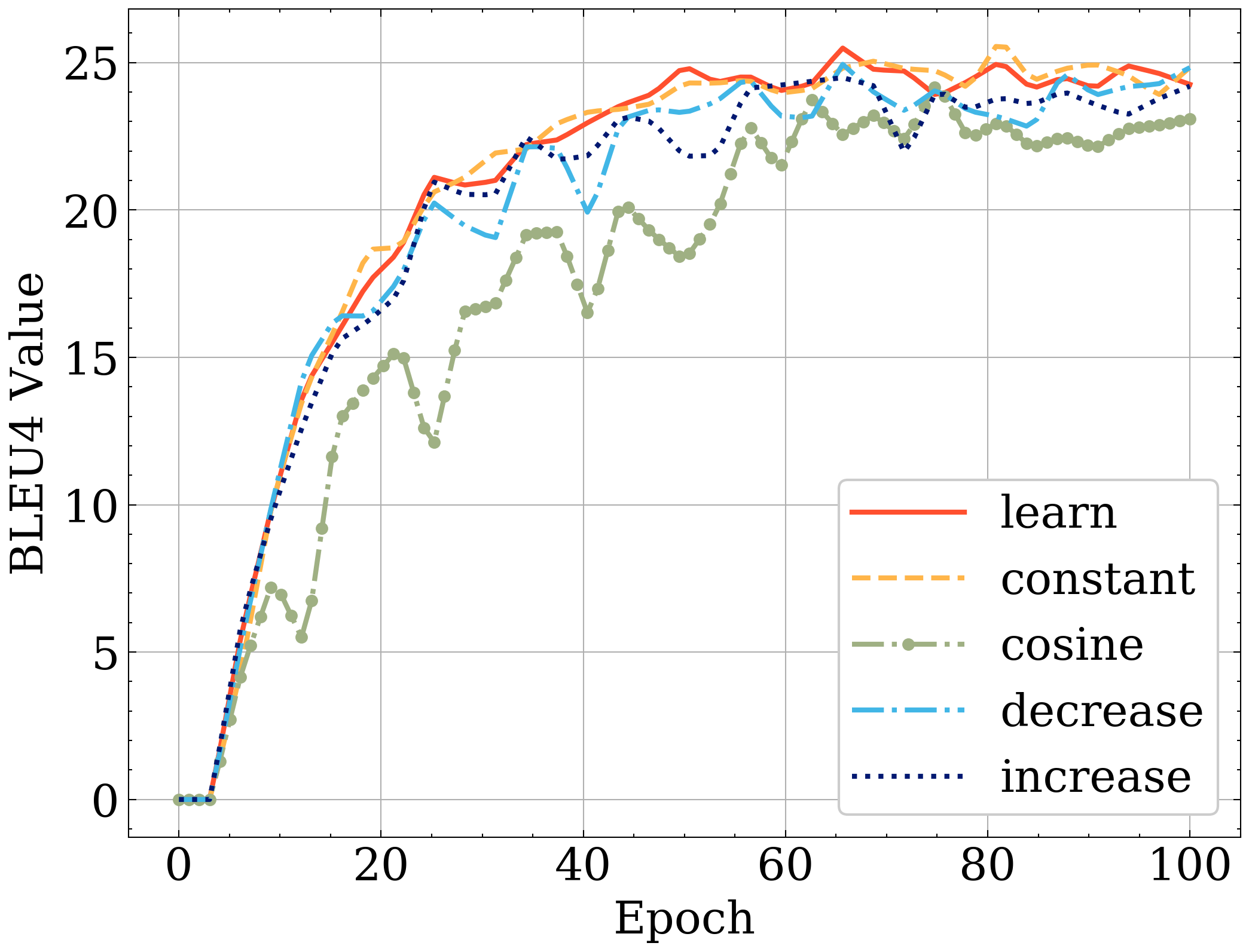}
		\label{fig:automl_subfigure5}}
	\quad
	\subfigure[Comparing various $\alpha$ strategies on ASLG dataset]{%
		\includegraphics[width=0.46\columnwidth]{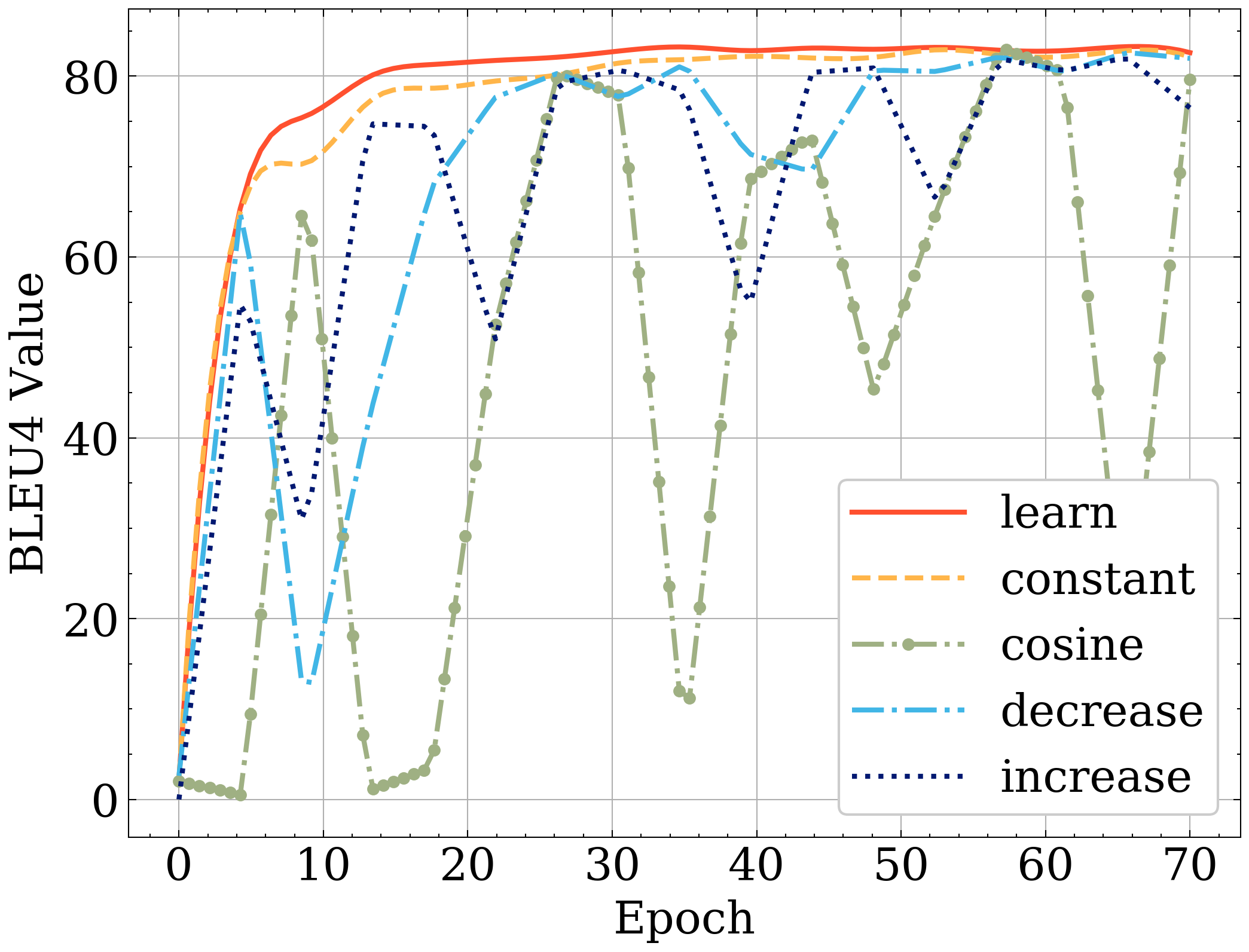}
		\label{fig:automl_subfigure6}}
	\quad
	\subfigure[The learned value of $\alpha$ on PH14 dataset]{%
		\includegraphics[width=0.46\columnwidth]{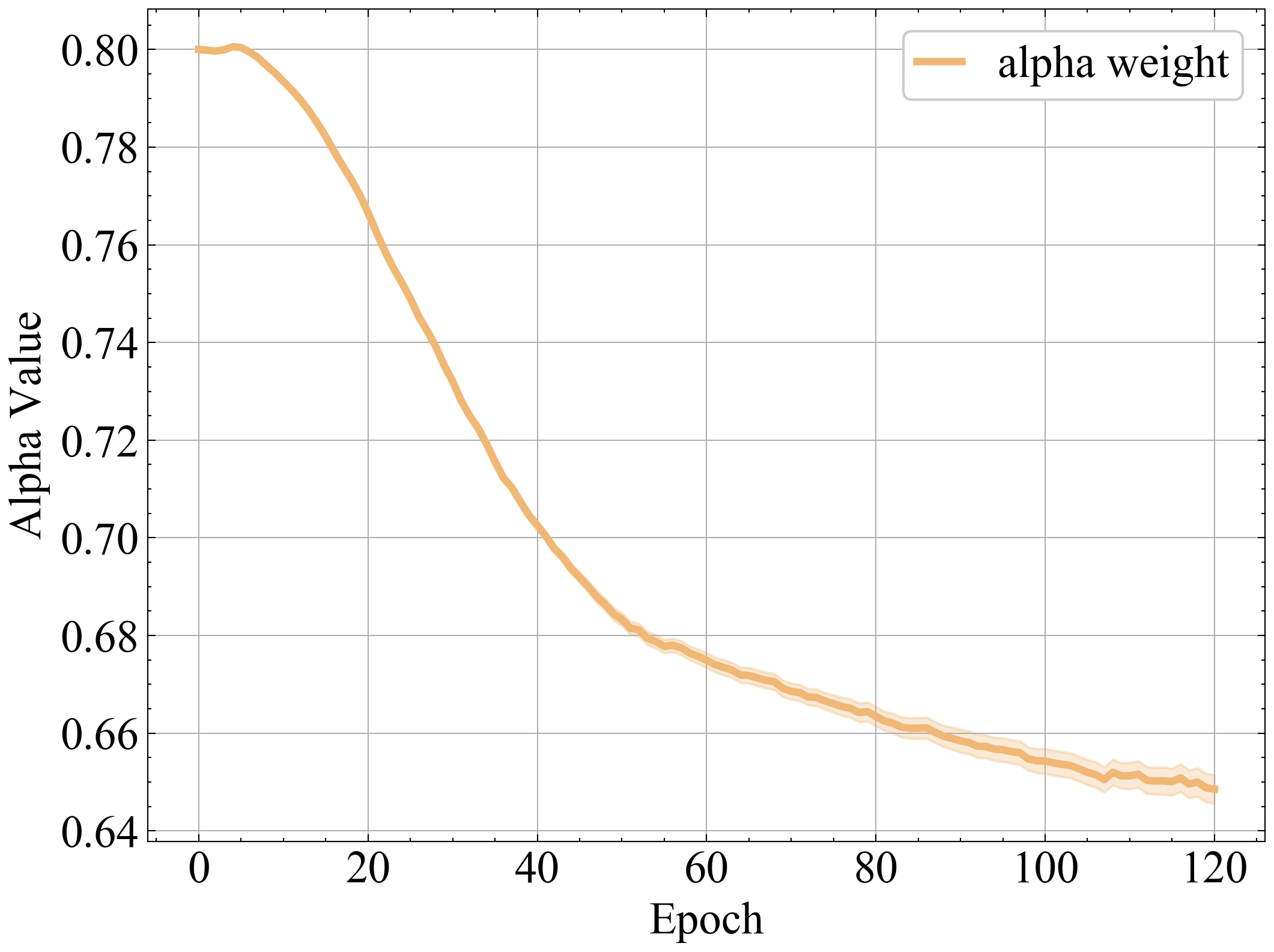}
		\label{fig:automl_subfigure7}}
	\quad
	\subfigure[The learned value of $\alpha$ on ASLG dataset]{%
		\includegraphics[width=0.46\columnwidth]{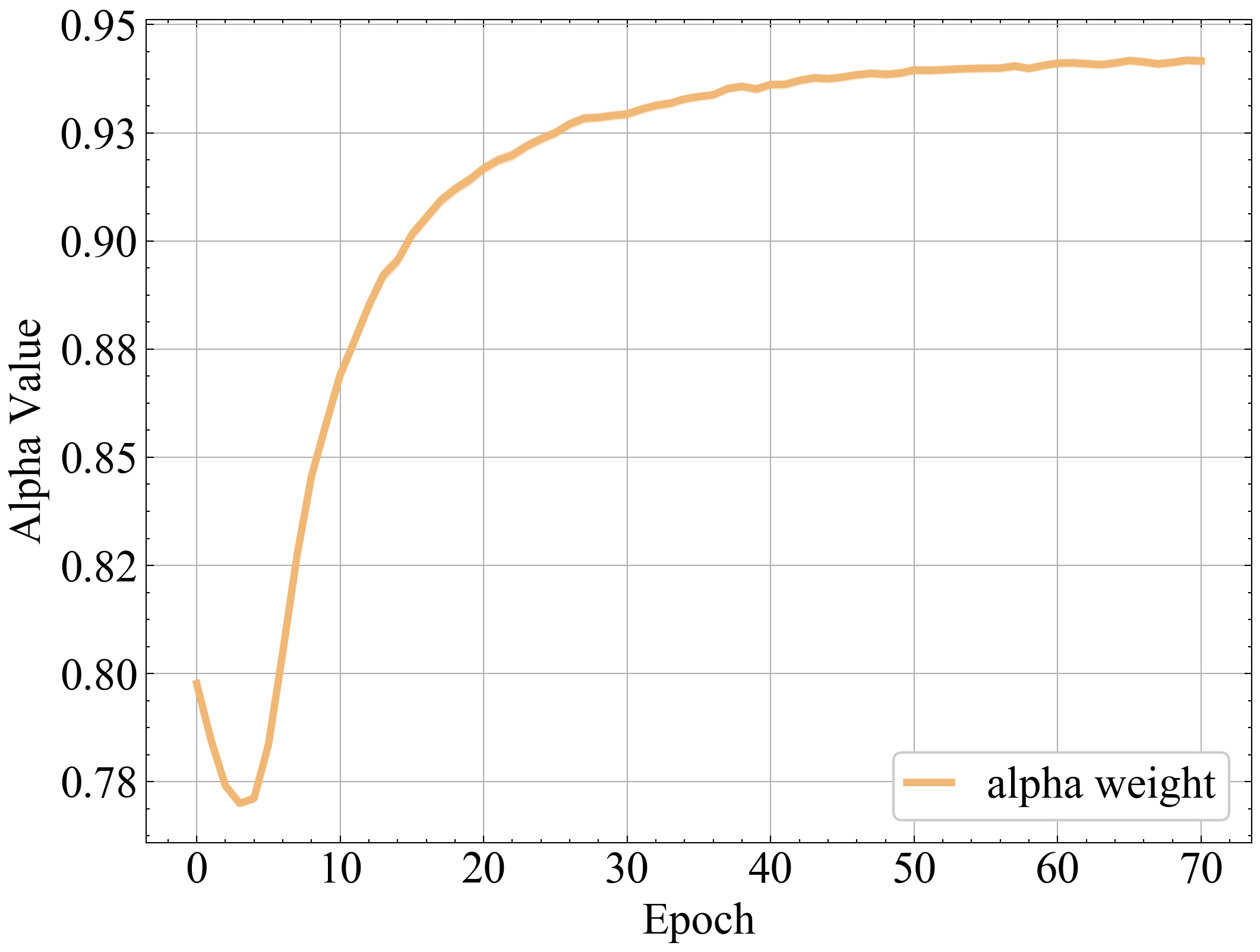}
		\label{fig:automl_subfigure8}}
	\quad
	\subfigure[Effect of beam size]{%
		\includegraphics[width=0.46\columnwidth]{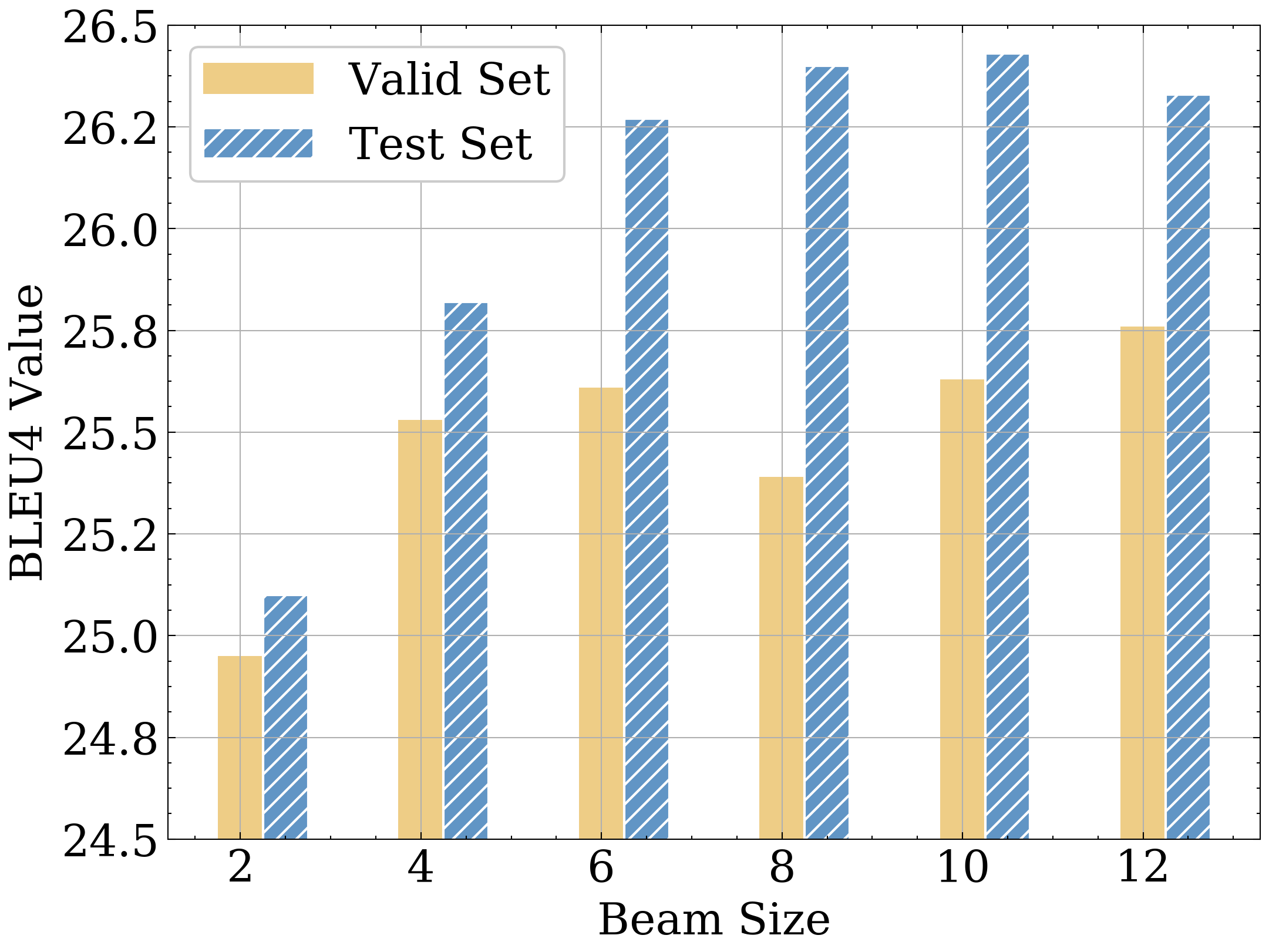}
		\label{fig:automl_subfigure1}}
	\quad
	\subfigure[Effect of layer number]{%
		\includegraphics[width=0.46\columnwidth]{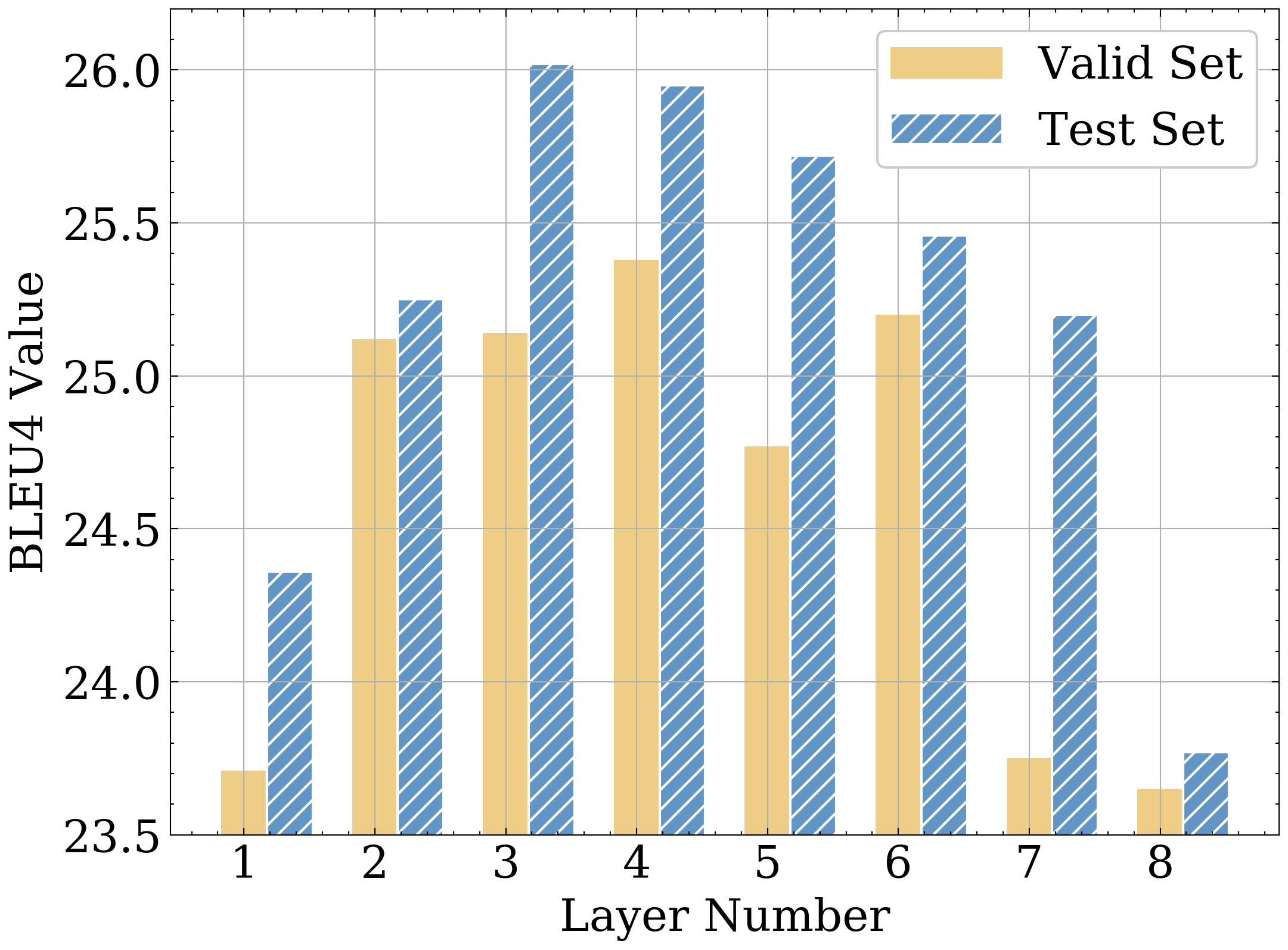}
		\label{fig:automl_subfigure2}}
	\quad
	\subfigure[Effect of learning rate]{%
		\includegraphics[width=0.46\columnwidth]{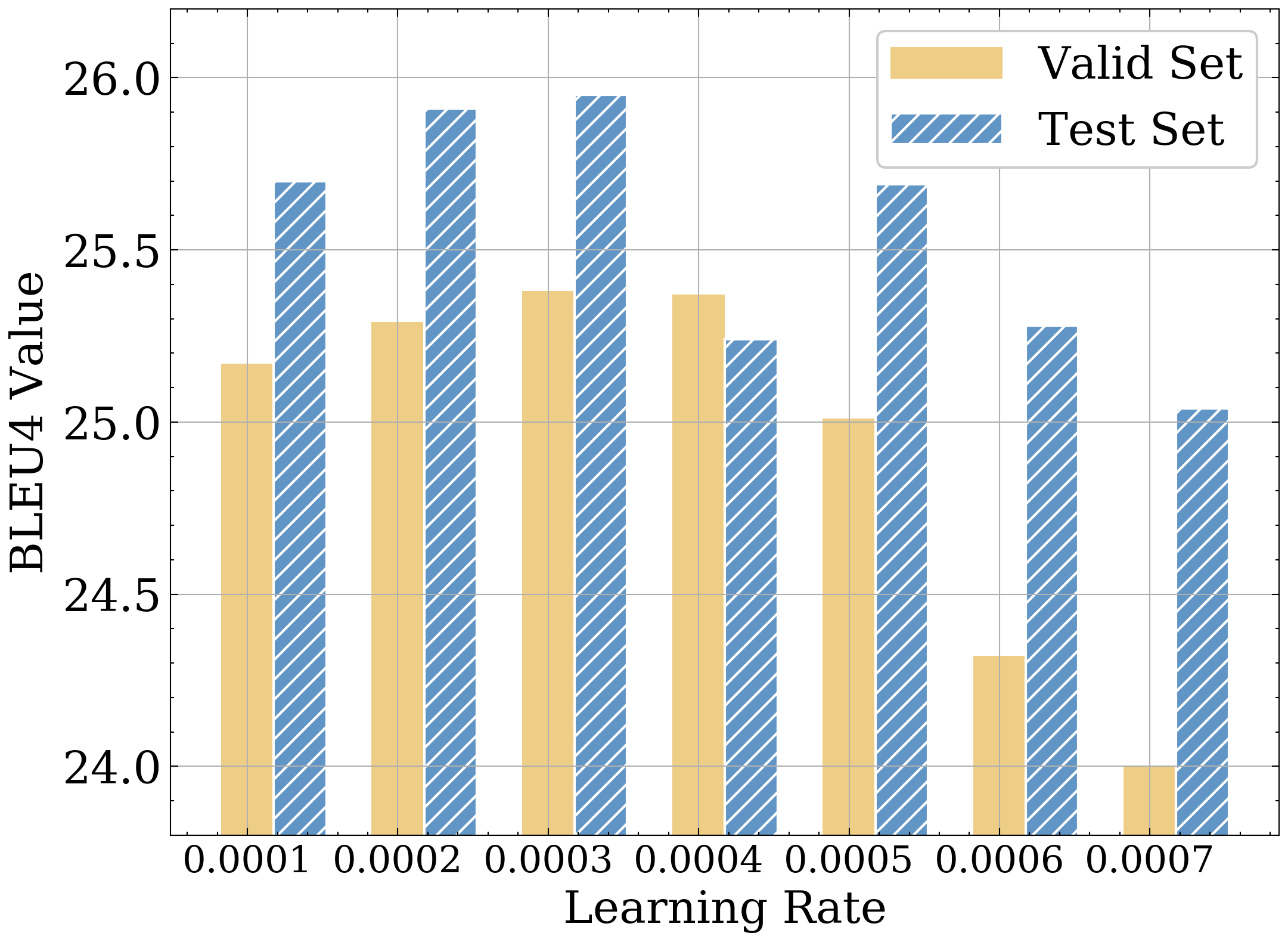}
		\label{fig:automl_subfigure3}}
	\quad
	\subfigure[Effect of dropout rate]{%
		\includegraphics[width=0.46\columnwidth]{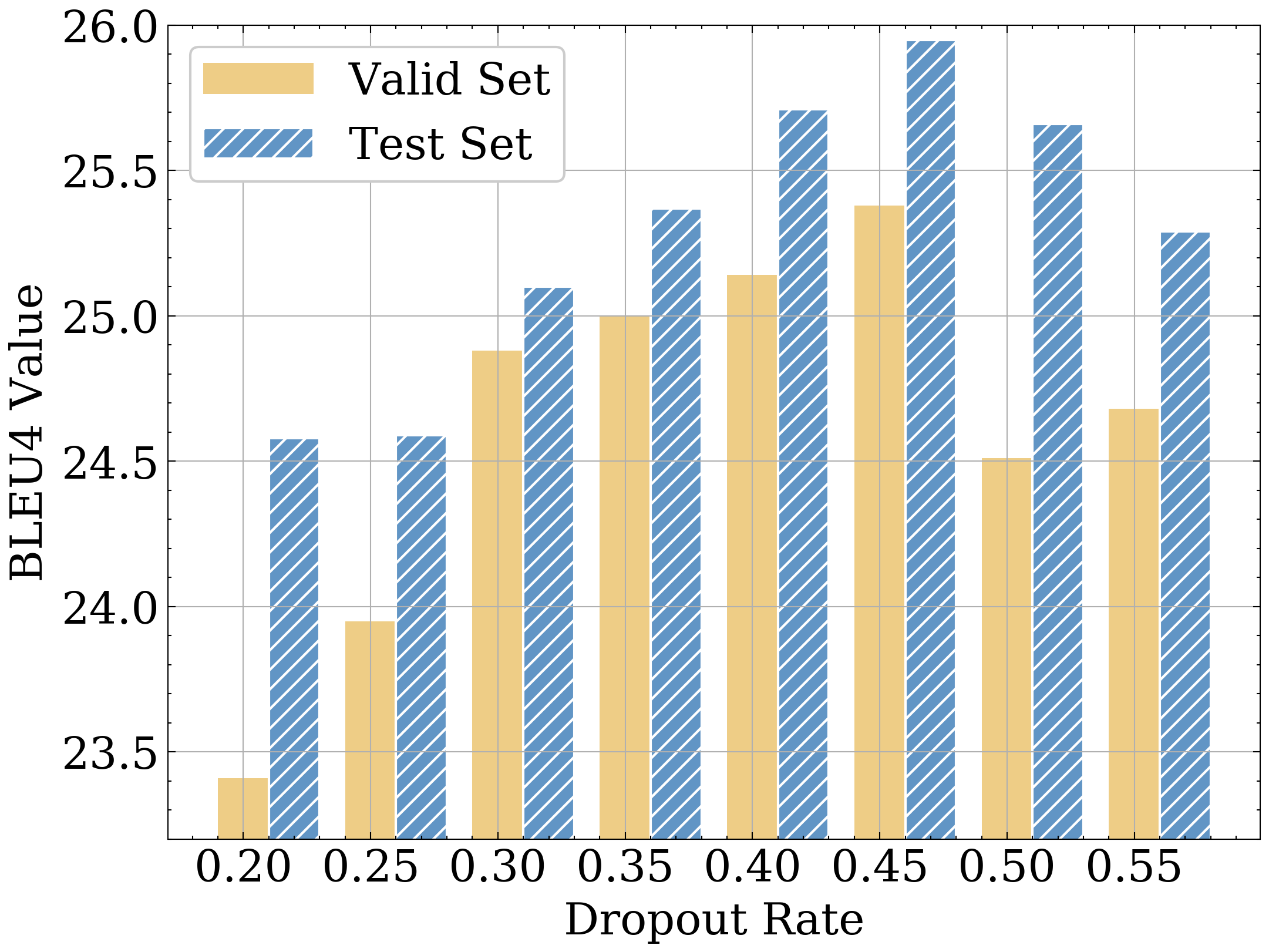}
		\label{fig:automl_subfigure4}}
	\caption{Various analysis results. (a) \& (b) present the results by using different feature fusion strategies on two datasets, respectively. (c) \& (d) show our learned value of $\alpha$ during the training process on the two datasets, respectively. (e)-(h) explore how beam size, layer number, learning rate, and dropout rate affect the model performance.}
	\label{fig:automl_result}
\end{figure*}

\para{Evaluation on S2G2T.} \
S2G2T is an extended task beyond G2T, which aims to recognize sign language videos to sign glosses, and then translate the recognized glosses to spoken sentences.
Hence, unlike the task of G2T, in this comparison, we focus on the evaluation of the whole two-step pipeline, that is, obtaining spoken language sentences from sign language videos.
Considering that only PH14 contains sign language videos, we thus conduct experiments on this dataset for S2G2T task, and the results are reported in Table \ref{result_S2G2T}.
Note that, for the recognition step, we employ STMC model to realize vision-based sequence learning \citep{zhou2020spatial}.
From the comparison we can see that, our TIN-SLT still outperforms existing approaches on most evaluation metrics.

\linespread{1.8}
\begin{table}[]
\begin{spacing}{1.35}
\resizebox{0.47\textwidth}{!}{
\begin{tabular}{l|llllll}
\Xhline{1pt}
\multicolumn{1}{c|}{}                                    & \multicolumn{6}{c}{Test Set}  \\ \cline{2-7}
\multicolumn{1}{c|}{\multirow{-2}{*}{\textbf{Model}}}    & \multicolumn{1}{c}{BLEU-1}    & \multicolumn{1}{c}{BLEU-2}    & \multicolumn{1}{c}{BLEU-3}
& \multicolumn{1}{c}{BLEU-4}                            & \multicolumn{1}{c}{ROUGL-L}   & METEOR                        \\ \hline \hline
G2T                 &  44.13  &  31.47  &  23.89  &  19.26  &  45.45  &  -      \\
S2G-G2T           &  41.54  &  29.52  &  22.24  &  17.79  &  43.45  &  -      \\
S2G2T             &  43.29  &  30.39  &  22.82  &  18.13  &  43.80  &  -      \\
Sign2     &  46.61  &  33.73  &  26.19  &  21.32  &  -      &  -      \\
Bahdanau            &  47.53  &  33.82  &  26.07  &  21.54  &  45.50  &  44.87  \\
Luong                &  47.08  &  33.93  &  26.31  &  21.75  &  45.66  &  44.84  \\
Transformer Ens.       &  50.63  &  38.36  &  30.58  &  25.40  &  \textbf{48.78}  &  47.60  \\ \hline
\textbf{TIN-SLT (Ours)}       & \textbf{51.06}    & \textbf{38.85}    & \textbf{31.23}
                                             & \textbf{26.13}     & 48.56 & \textbf{47.83}      \\
\Xhline{1pt}
\end{tabular}}
\caption{\label{result_S2G2T} Comparing the S2G2T performance by using our TIN-SLT and state-of-the-art techniques on PHOENIX-2014-T dataset. The results of G2T, S2G-G2T and S2G2T are from \citep{camgoz2018neural}. The results of Sign2 are from \citep{camgoz2020sign}. The results of Bahdanau, Luong, and Transformer Ens. are from \citep{yin2020better}. Clearly, our TIN-SLT achieves the highest values on most metrics.}
\end{spacing}
\vspace{-2mm}
\end{table}

\subsection{Analysis and Discussions}
Here, we conducted a series of detailed experiments to analyze our method and give some insights behind our network design.

\para{Effect of learning-based feature fusion.} \
In this work, we propose a learning-based strategy to set $\alpha$ dynamically.
Here, we conducted experiments by comparing this strategy with other four different strategies, including (1) cosine annealing \citep{loshchilov2016sgdr}, (2) cosine increment, (3) cosine decrement, and (4) constant value.
The update of $\alpha$ by the three cosine strategies are calculated as Eq. (\ref{equ:cosine}) with different settings of the epoch cycle coefficient $T_c$:
\begin{equation}\label{equ:cosine}
	\alpha_{t+1} = \alpha_{min} + \frac{1}{2} (\alpha_{max} - \alpha_{min})(1-cos(\frac{T_{t}}{T_c}\pi + \gamma))
	\label{alpha}
\end{equation}
where $\alpha$ is the fusion ratio, $T_{t}$ is current epoch step, and $\gamma$ is the time-shift constant. We set $T_c$ as (25, 100, 100) and $\gamma$ as (0, 0, $\pi$) for cosine annealing, cosine decrement, and cosine increment, respectively.
The minimum value $\alpha_{min}$ and maximum value $\alpha_{max}$ of $\alpha$ are set to be 0 and 1.

Figures \ref{fig:automl_subfigure5}-\ref{fig:automl_subfigure6} are the experimental results on the two datasets.
We can observe that the learning-based strategy (red line) gets the best result on ASLG and comparable result with the constant setting ($\alpha$$=$$0.8$) on PH14, but still better than other three cosine strategies.
Moreover, we also visualize the learned value of $\alpha$ during the training process as shown in Figures \ref{fig:automl_subfigure7}-\ref{fig:automl_subfigure8} to find out the contribution ratio of the BERT model to the final performance.
We can see that, the value of $\alpha$ is gradually decreasing on PH14, meaning that the model depends more on the BERT pre-trained knowledge at the beginning of the training process and gradually inclines to our employed training corpus. The observation is just opposite on ASLG, since it is a much larger dataset than PH14 and our model relies more on BERT to further boost the performance near the end of training.

\linespread{1.8}
\begin{table}[t]
	\begin{spacing}{1.35}
		\resizebox{0.467\textwidth}{!}{
			\begin{tabular}{l|llllll}
				\Xhline{1pt}
				\multicolumn{1}{c|}{}                                    & \multicolumn{6}{c}{Test Set}  \\ \cline{2-7}
				\multicolumn{1}{c|}{\multirow{-2}{*}{\textbf{Model}}}    & \multicolumn{1}{c}{BLEU-1}    & \multicolumn{1}{c}{BLEU-2}    & \multicolumn{1}{c}{BLEU-3}
				& \multicolumn{1}{c}{BLEU-4}                            & \multicolumn{1}{c}{ROUGE-L}   & METEOR                        \\ \hline \hline
				&\multicolumn{6}{c}{\textbf{PHOENIX-2014-T Dataset Evaluation}}  \\
				Baseline            &  47.69  &  35.52  &  28.17  &  23.32  &  46.58  &  44.85   \\
				w/ DataAug          &  50.77 & 37.85 & 29.88 & 24.57 &  47.39 &  46.95  \\
				w/ Encoder          &  51.05 & 37.94 & 29.91 & 24.63 & 47.59 & 47.13    \\
				w/ Decoder           &  50.99 & 38.47 & 30.48 & 25.08 & 48.78 & 48.20    \\
				\textbf{Full pipeline}                     &  \textbf{52.77}   & \textbf{40.08}    & \textbf{32.09}    & \textbf{26.55}    & \textbf{49.43}    & \textbf{49.36}         \\  \hline
				&\multicolumn{6}{c}{\textbf{ASLG-PC12 Dataset Evaluation}}  \\
				Baseline            &  92.98  &  89.09  &  85.63  &  82.41  &  \textbf{95.87}   &  96.46           \\
				w/ DataAug          & 92.60 &  89.15 &  85.80 &  83.05 &  95.08 &  95.33  \\
				w/ Encoder         &  92.77 &  89.22 &  86.23 &  83.40 &  95.22 &  \textbf{96.87} \\
				w/ Decoder           & 93.15 &  89.80 &  86.49 &  83.89 &  95.34 &  95.67   \\
				\textbf{Full pipeline}         &  \textbf{93.35}  &  \textbf{90.03}  &  \textbf{87.07}  &  \textbf{84.29}
				& 95.39   & 95.92   \\
				\Xhline{1pt}
		\end{tabular}}
		\caption{\label{result_ablation_compare} Ablation analysis of our major network components on the G2T task.
		}
	\end{spacing}
\end{table}

\para{Analysis on major network components.} \
In our TIN-SLT, there are two major components: the multi-level data augmentation scheme and the instruction module.
To validate the effectiveness of each component, we conduct an ablation analysis on the G2T task with the following cases.

\begin{itemize}
	\vspace*{-2mm}
	\item Baseline: We use two layers Transformer \citep{yin2020better} without data augmentation and instruction module as baseline.
	\vspace*{-2mm}
	\item w/ DataAug: Based on the baseline, we add our data augmentation scheme back.
	\vspace*{-2mm}
	\item w/ Encoder: Based on w/ DataAug, we fuse instruction module only into the encoder.
	\vspace*{-2mm}
	\item w/ Decoder: Based on w/ DataAug, we fuse instruction module only into the decoder.
	\vspace*{-2mm}
\end{itemize}
As a contrast, in our full pipeline, the instruction module is inserted into both encoder and decoder.
Table \ref{result_ablation_compare} shows the evaluation results on both PH14 and ASLG.
By comparing the results from Baseline and w/ DataAug, we can see that our data augmentation improves the translation performance, especially for the PH14 dataset.
A reasonable interpretation is that the translation task on PH14 dataset is more difficult than on ASLG, thus our data augmentation contributes more.
On the other hand, w/ Encoder, w/ Decoder and Full pipeline explore the best location to introduce PTM information into the model.
Results in Table \ref{result_ablation_compare} show that our full model achieves the best performance.
Particularly, by comparing the results from w/ Encoder and w/ Decoder against the results from SOTA methods (Tables~\ref{G2T_result} \& \ref{result_ablation_compare}), we can observe that as long as we employ the pre-trained model, no matter where it is inserted into the network, the performance is always better than existing methods.

\begin{table}[]
\begin{spacing}{1.35}
\resizebox{0.467\textwidth}{!}{
\begin{tabular}{l|lllll}
\Xhline{1pt}
Model\protect\footnotemark[1]   &  Size(MB)  &  Dataset  &  Gloss(\%)  &  Text(\%)  &  BLEU4           \\ \hline \hline
&\multicolumn{5}{c}{\textbf{PHOENIX-2014-T Dataset Evaluation}}  \\
Multilingual                    &  641.10    &   PH14    &  59.96      &  74.62     &  25.48           \\
Distilbert                      &  257.30    &   PH14    &  44.50      &  71.15     &  24.73           \\
Gbert                           &  421.80    &   PH14    &  44.50      &  71.15     &  25.13           \\
Dbmdz                           &  421.80    &   PH14    &  73.72      &  88.13     &  \textbf{26.55}  \\  \hline
&\multicolumn{5}{c}{\textbf{ASLG-PC12 Dataset Evaluation}}  \\
Base-Tiny                       & 16.90      &   ASLG    &  76.77      &  96.35     &  82.44           \\
Electra                         &  51.70     &   ASLG    &  76.77      & 96.35      &  82.60            \\
Distilbert                      & 255.60     &   ASLG    &  76.77      & 96.35      &  83.06            \\
Base-uncased                            & 420.10     &   ASLG    &  76.77      & 96.35      &  \textbf{84.29}  \\
 \Xhline{1pt}
\end{tabular}}
\end{spacing}
\vspace*{-2mm}
\caption{\label{result_bert} Comparing different pre-trained models in terms of BLEU-4.
}
\end{table}
\vspace{-1mm}
\footnotetext[1]{The pre-trained models links are listed in Appendix.}

\para{Effect of different pre-trained models.} \
We here explored the translation performance by using different pre-trained models; see Table \ref{result_bert}.
We analyzed the model size and vocabulary coverage of the pre-trained model with gloss and text of our dataset.
We can see that introducing a pre-trained model with larger vocabulary coverage of the target dataset will gain better performance, since a pre-trained model with larger vocabulary coverage can inject more knowledge learned from another unlabeled corpus into the translation task.
For ASLG, although the vocabulary coverage is the same, we can see that the bigger model has better performance since it can learn contextual representation better.

\para{Analysis on hyper-parameters.} \
To search the best settings of our hyper-parameters, we employed Neural Network Intelligence (NNI) \citep{nni2018}, a lightweight but powerful toolkit.
As shown in Figures \ref{fig:automl_subfigure1}-\ref{fig:automl_subfigure4},
we explored how beam size, layer number, learning rate and dropout rate affect the model performance on PH14 dataset.
First,
beam search enables to explore more possible candidates, but large beam widths do not always result in better performance as shown in Figure \ref{fig:automl_subfigure1}. We obtain optimal beam size as 10 on PH14.
Second,
the layer number decides the model size and capacity, where the larger model would overfit on a small dataset. In Figure \ref{fig:automl_subfigure2}, we find the optimal layer number to be 3 on PH14.
Lastly, as shown in Figures \ref{fig:automl_subfigure3} \& \ref{fig:automl_subfigure4}, we adopt an early-stopping strategy to avoid overfitting and find the best learning rate and dropout rate are 0.0003 and 0.45, respectively.

\para{Case study.} \
Table \ref{example_aslg} presents some intuitive translation results on ASLG by reporting the translated spoken sentences.
Overall, the translation quality is good, even the translated sentences with low BLEU-4 still convey the same information.
Also, we can observe that our translated sentences are basically the same with ground truth, although using different expressions, i.e., ``decision making'' vs. ``decision made''.
The translation results on PH14 are reported in Appendix.

\begin{table}[t]
\begin{spacing}{1.1}
\resizebox{0.47\textwidth}{!}{
\begin{tabular}{l|l|l}
\Xhline{1pt}
Type                       & Content                                                          & BLEU-4                  \\ \hline \hline
GT Gloss                   & X-IT BE DESC-UP TO X-YOU TO CONSIDER                             & \multirow{6}{*}{100.00} \\
                           & AND CHOOSE OUTCOME X-YOU WANT TO SEE .                           &                         \\
GT Text                    & it is up to you to consider and choose                           &                         \\
                           & the outcome you want to see .                                    &                         \\
Pred Text                  & it is up to you to consider and choose                           &                         \\
                           & the outcome you want to see .                                    &                         \\ \hline
GT Gloss                   & X-I WANT IRELAND TO REMAIN AT                                    & \multirow{6}{*}{57.58}  \\
                           & HEART DECISION MAKE IN EUROPE .                                  &                         \\
GT Text                    & i want ireland to remain at the                                  &                         \\
                           & heart of decision making in europe .                             &                         \\
Pred Text                  & i want ireland to remain at the                                  &                         \\
                           & heart of the decision made in europe .                           &                         \\ \hline
GT Gloss                   & X-I WILL DESC-NEVER FORGET WHAT X-I                              & \multirow{5}{*}{13.44}  \\
                           & EXPERIENCE . SHOULD BE ABOUT .                                   &                         \\
GT Text                    & that is what this european day of memorial should be             &                         \\
                           & about . i will never forget what i experienced .                 &                         \\
Pred Text                  & i will never forget what i experienced .                         &                          \\ \Xhline{1pt}
\end{tabular}}
\caption{\label{example_aslg} Qualitative evaluation of translation performance in different BLEU-4 scores on ASLG dataset.}
\end{spacing}
\vspace{-1mm}
\end{table}

\vspace{-2mm}

\section{Conclusion}\label{sec:Conclusion}

In this paper, we proposed a task-aware instruction network for sign language translation. To address the problem of limited data for SLT, we introduced a pre-trained model into Transformer and designed an instruction module to adapt SLT task. Besides, due to the discrepancy between the representation space of sign glosses and spoken sentences, we proposed a multi-level data augmentation scheme. Extensive experiments validate our superior performance compared with state-of-the-art approaches.
While there is obvious improvement among most evaluation metrics, the complexity of our models is also increased, causing a longer training period.
In the future, we would like to explore the possibility of designing a lightweight model to achieve real-time efficiency.

\section*{Acknowledgements}

We thank anonymous reviewers for the valuable comments. This work is supported by the China National Natural Science Foundation (No. 62176101 \& No. 62106094) and Zhejiang Lab's International Talent Fund for Young Professionals.

\bibliography{anthology,custom}
\clearpage
\appendix


\section{Appendix}

\subsection{Dataset Description}
In this section, we will introduce two public benchmark datasets used in sign language translation tasks, namely PHOENIX-2014-T and ASLG-PC12. We conducted statistical analysis on the datasets and the results are shown in Table \ref{dataset-table}. It is obvious that PHOENIX-2014-T is a small-scale dataset, while ASLG-PC12 is a large-scale dataset.
\begin{table}[h]
\centering
\begin{spacing}{1.5}
\resizebox{\columnwidth}{!}{
\begin{tabular}{llllllll}
\Xhline{1pt}
\multicolumn{2}{l}{\multirow{2}{*}{Dataset}}        & \multicolumn{3}{c}{Gloss}   & \multicolumn{3}{c}{Translation} \\ \cline{3-8}
\multicolumn{2}{l}{}                                & Train   & Dev    & Test     & Train     & Dev      & Test          \\ \hline \hline
\multirow{3}{*}{\textbf{PH14}}      & Samples        & 7096    & 519    & 642      & 7096      & 519      & 642           \\
                                    & Vocabs         & 1066    & 393    & 411      & 2887      & 951      & 1001          \\
                                    & Words         & 67781   & 3745   & 4257     & 99081     & 6820     & 7816          \\ \hline
\multirow{3}{*}{\textbf{ASLG}}      & Samples        & 82709   & 4000   & 1000     & 82709     & 4000     & 1000          \\
                                    & Vocabs         & 15782   & 4323   & 2150     & 21600     & 5634     & 2609          \\
                                    & Words         & 862046  & 41030  & 10503    & 975942    & 46637    & 11953         \\ \Xhline{1pt}
\end{tabular}}
\end{spacing}
\caption{\label{dataset-table} The descriptive statistics of PHOENIX-2014-T and ASLG-PC12 datasets. Samples row means the sample size of the dataset, Vocabs row represents the total vocabularies contained in the dataset, and Words row means the total words of the dataset. }
\end{table}

\subsection{PHOENIX-2014-T Qulitative Result}

BE-SLT performance of G2T task on PHOENIX-2014-T is shown in Table \ref{example_ph14}, from which we can observe that sign language translation results are of good quality with different BLEU-4 scores and the predicted sentences can convey effective information even for low BLEU-4 scores.

\begin{table}[h]
\begin{spacing}{1.5}
\resizebox{0.46\textwidth}{!}{
\begin{tabular}{l|l|l}
\Xhline{1pt}
Type                       & Content                                                                             & BLEU-4                  \\ \hline \hline
Gloss                      & BERG ORKAN MOEGLICH                                                                 & \multirow{5}{*}{100.00} \\
GT Text                    & auf den bergen sind orkanartige                                                     &                         \\
                           &   b{\"o}en m{\"o}glich .                        &                         \\
Pred Text                  &  auf den bergen sind orkanartige                                                 &                         \\
                           &    b{\"o}en m{\"o}glich .                       &                         \\ \hline
Gloss                      & HEUTE NACHT ZWISCHEN NEUNZEHN ZWISCHEN                                              & \multirow{6}{*}{57.58}  \\
                           & FUENFZEHN SUEDOST MAXIMAL ZWOELF                                                    &                         \\
GT Text                    & heute nacht werte zwischen neunzehn und f{\"u}nfzehn               &                         \\
                           & grad im s{\"u}dosten bis zw{\"o}lf grad .         &                         \\
Pred Text                  & heute nacht neunzehn bis f{\"u}nfzehn grad im                                     &                         \\
                           &s{\"u}dosten bis zw{\"o}lf grad .                 &                         \\ \hline
Gloss                      & RUSSLAND IX TROCKEN HEISS SCHEINEN FUENF                                            & \multirow{6}{*}{13.44}  \\
                           & DREISSIG BIS VIERZIG GRAD                                                           &                         \\
GT Text                    & ganz anders die trockene hitze {\"u}ber russland                   &                         \\
                           & mit f{\"u}nfunddrei{\ss}ig bis vierzig grad .    &                         \\
Pred Text                  & aber bei uns wird es auch noch ein bisschen                                       &                         \\
                           & hei{\ss}er da sind es f{\"u}nf bis vierzig grad . &                         \\ \Xhline{1pt}
\end{tabular}}
\caption{\label{example_ph14} PHOENIX-2014-T: Qualitatively evaluation of translation performance in different BLEU-4 scores. }
\end{spacing}
\end{table}

\subsection{Experiment Parameter}
In order to help reproduce BE-SLT and its translation performance, as shown in Table \ref{para_ph14} and \ref{para_aslg}, we list the hyper-parameters of the best results on two benchmark datasets.
For G2T task on PHOENIX-2014-T, we list the best hyper-parameter settings for the experiments which apply data augmentation scheme, or fuse BERT-attention module into encoder, decoder, and both respectively (namely,w/DataAug, w/Encoder, w/Decoder, w/All). W/All obtains the highest BLEU-4 using the initial learning rate of 0.00025, dropout rate of 0.45, beam search with width 5, and the max epoch size of 120.
For G2T task on ASLG-PC12, we also list the hyper-parameter settings for the four experiments that achieve significant results, listed in Table \ref{para_aslg}. For more experiment details, please refer to our code which will be published upon the publication of this work.

\linespread{1.8}
\begin{table}[htbp]
\centering
\begin{spacing}{1.35}
\resizebox{0.467\textwidth}{!}{
\begin{tabular}{c|cccc}
\Xhline{1pt}
\multicolumn{1}{c}{}                                    & \multicolumn{4}{c}{PHOENIX-2014-T}  \\  \cmidrule{2-5}
\multicolumn{1}{c}{\multirow{-2}{*}{\textbf{Parameter}}}    &
\multicolumn{1}{c}{w/DataAug} & \multicolumn{1}{c}{w/Encoder} &
\multicolumn{1}{c}{w/Decoder} &
\multicolumn{1}{c}{w/All}
\\
\hline \hline
Embedding size & 512 &512 &512 &512  \\
Hidden size & 2048  & 2048 & 2048 & 2048 \\
Head number   & 8 & 8 & 8 & 8 \\
Encoder BERT gate   & 1  & 1 & 0 & 1 \\
Decoder BERT gate  & 1  & 0 & 1 & 1 \\
\hline
Optimizer   & Adam & Adam   & Adam & Adam\\
Learning rate    &  0.00025 &  0.00025  &  0.00025  &  0.0003\\
LR schedule   & inverse sqrt & inverse sqrt & inverse sqrt & inverse sqrt \\
Weight decay & $10^{-3}$ & $10^{-3}$ & $10^{-3}$ & $10^{-3}$ \\
Drop out  &  0.45  &  0.45  &  0.45  &  0.45      \\
Label smoothing      &  0.3 &  0.3  &  0.3  &  0.3 \\
BERT ratio  & - & 0.6 & 0.6  & 0.65 \\
Max epoch & 120 & 120 & 120 & 120 \\
BERT model  & \multicolumn{4}{c}{ bert-base-german-dbmdz-uncased} \\
\Xhline{1pt}
\end{tabular}}
\caption{\label{para_ph14} The hyper-parameters of the best results on PHOENIX-2014-T for the G2T task.}
\end{spacing}
\end{table}

\linespread{1.8}
\begin{table}[htbp]
\centering
\begin{spacing}{1.35}
\resizebox{0.467\textwidth}{!}{
\begin{tabular}{c|cccc}
\Xhline{1pt}
\multicolumn{1}{c}{}                                    & \multicolumn{4}{c}{ASLG-PC12}  \\  \cmidrule{2-5}
\multicolumn{1}{c}{\multirow{-2}{*}{\textbf{Parameter}}}    &
\multicolumn{1}{c}{w/DataAug} & \multicolumn{1}{c}{w/Encoder} &
\multicolumn{1}{c}{w/Decoder} &
\multicolumn{1}{c}{w/All}
\\
\hline \hline
Embedding size & 512 &512 &512 &512  \\
Hidden size & 2048  & 2048 & 2048 & 2048 \\
Head number   & 8 & 8 & 8 & 8 \\
Encoder BERT gate   & 1  & 1 & 0 & 1 \\
Decoder BERT gate  & 1  & 0 & 1 & 1 \\
\hline
Optimizer   & Adam & Adam   & Adam & Adam\\
Learning rate    &  0.00025 &  0.00025  &  0.00025  &  0.00045\\
LR schedule   & inverse sqrt & inverse sqrt & inverse sqrt & inverse sqrt \\
Weight decay & $10^{-3}$ & $10^{-3}$ & $10^{-3}$ & $10^{-3}$ \\
Drop out  &  0.45  &  0.45  &  0.45  &  0.4      \\
Label smoothing      &  0.3 &  0.3  &  0.3  &  0.1 \\
BERT ratio  & -  & 0.6 & 0.6  & 0.6 \\
Max epoch & 70 & 70 & 70 & 70 \\
BERT model  & \multicolumn{4}{c}{ bert-base-uncased} \\
\Xhline{1pt}
%
\end{tabular}}
\caption{\label{para_aslg} The hyper-parameters of the best results on ASLG-PC12 for the G2T task.}
\end{spacing}
\end{table}

\subsection{Alpha Strategy Settings}
\begin{figure}[h]
	\centering
	\subfigure[Cosine annealing strategy]{%
		\includegraphics[width=0.48\columnwidth]{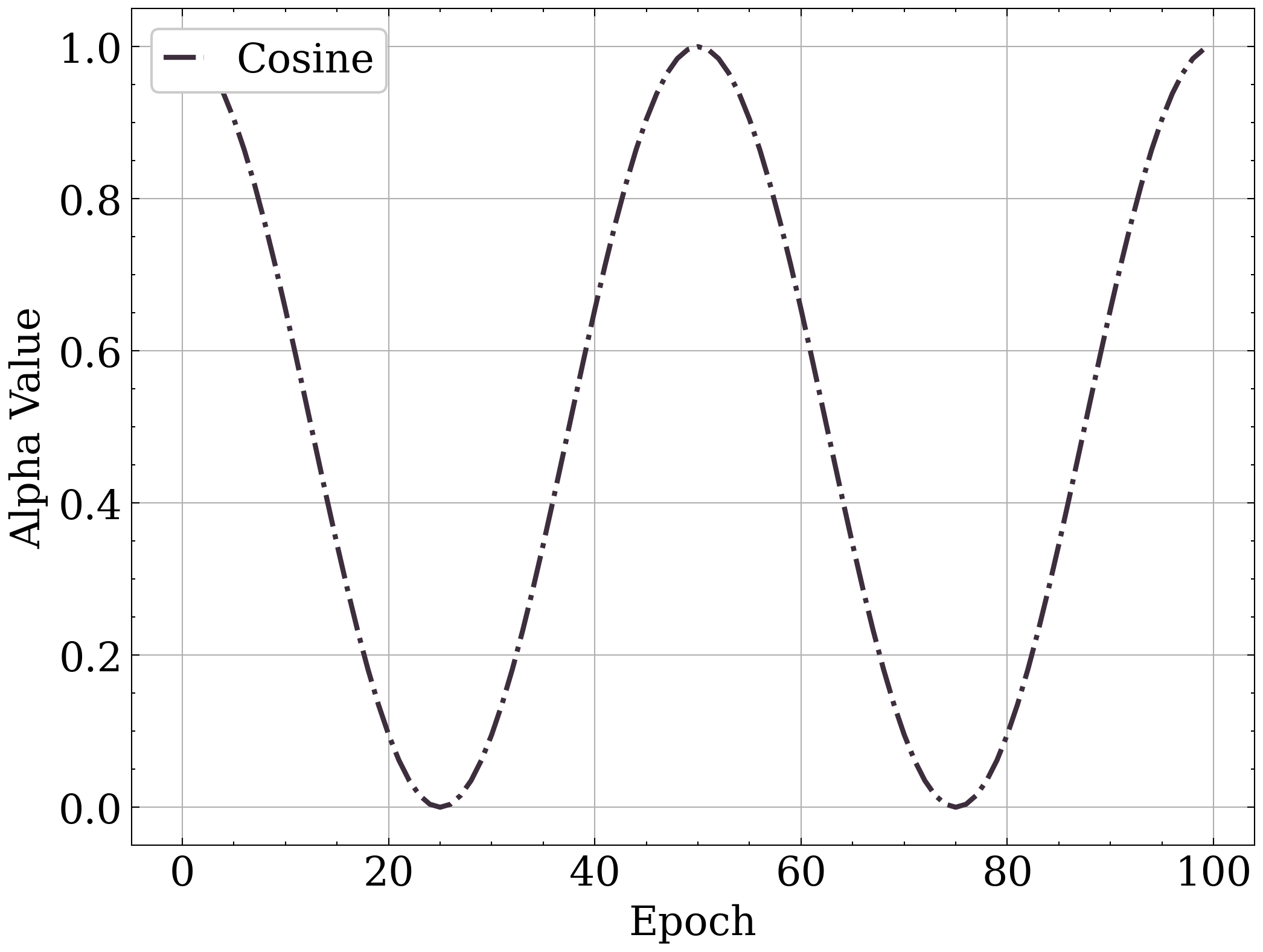}
		\label{fig:subfigure1}}
	\subfigure[Constant strategy]{%
		\includegraphics[width=0.48\columnwidth]{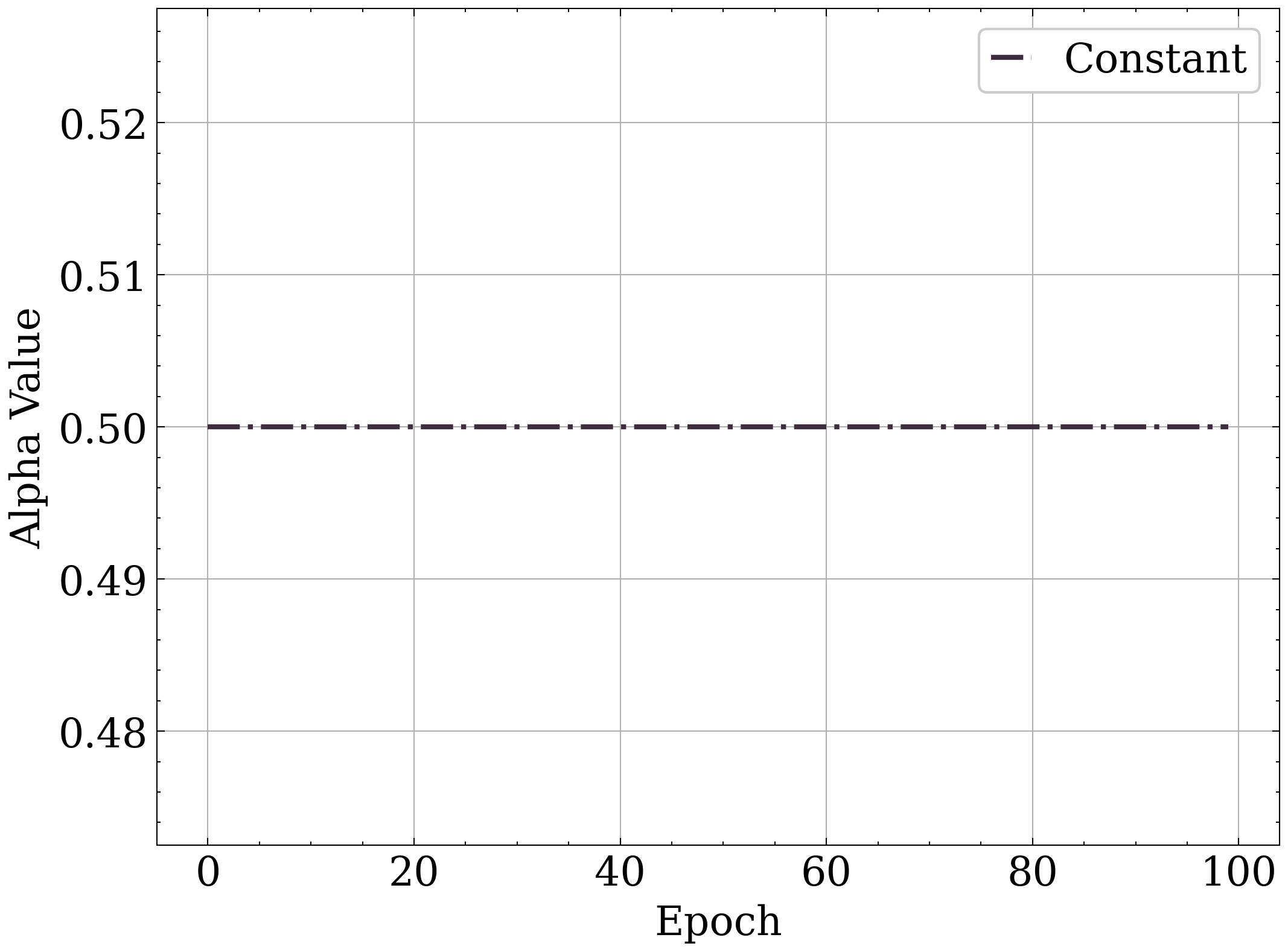}
		\label{fig:subfigure2}}
	\subfigure[Cosine increment strategy]{%
		\includegraphics[width=0.48\columnwidth]{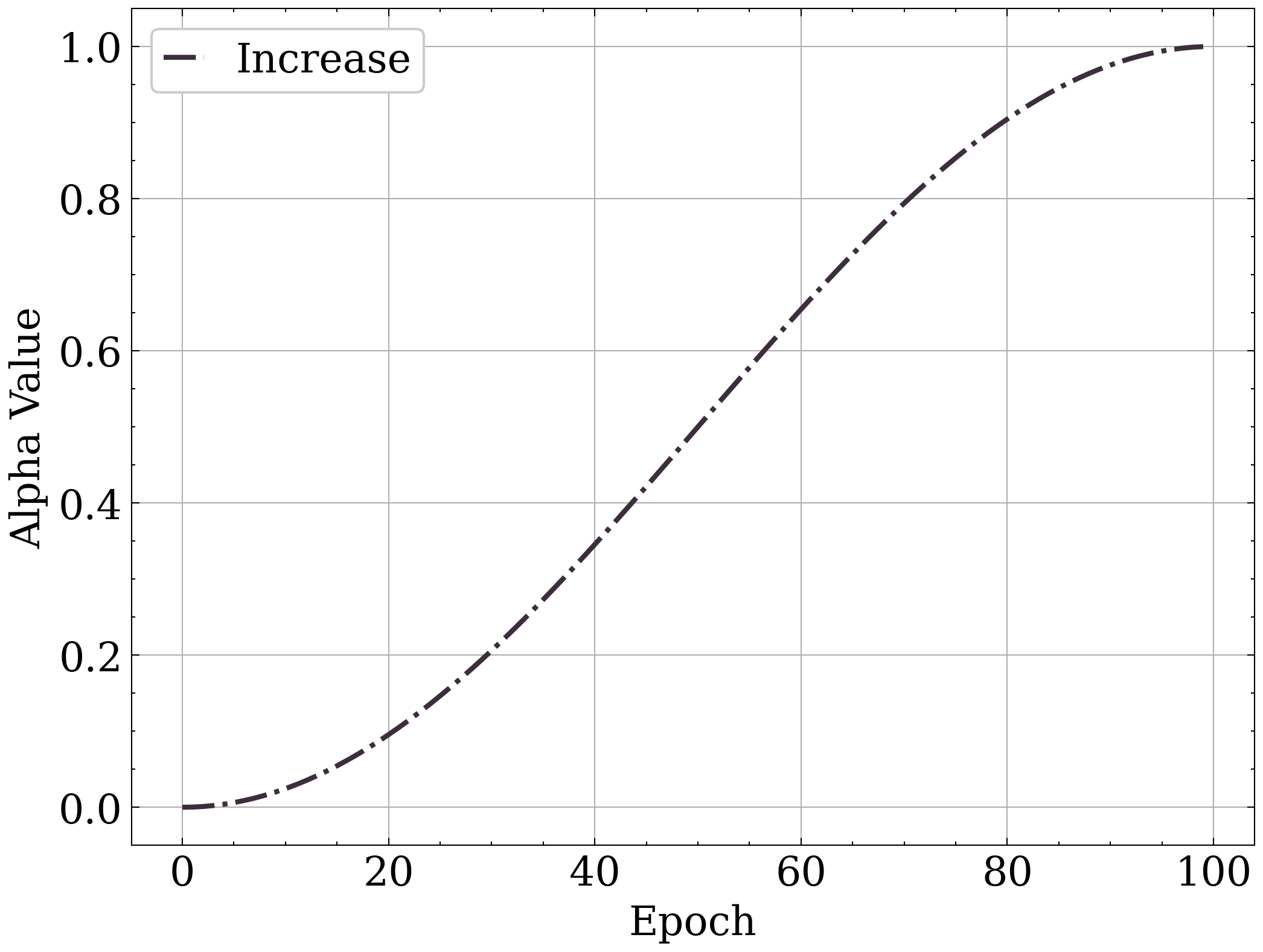}
		\label{fig:subfigure1}}
	\subfigure[Cosine decrement strategy]{%
		\includegraphics[width=0.48\columnwidth]{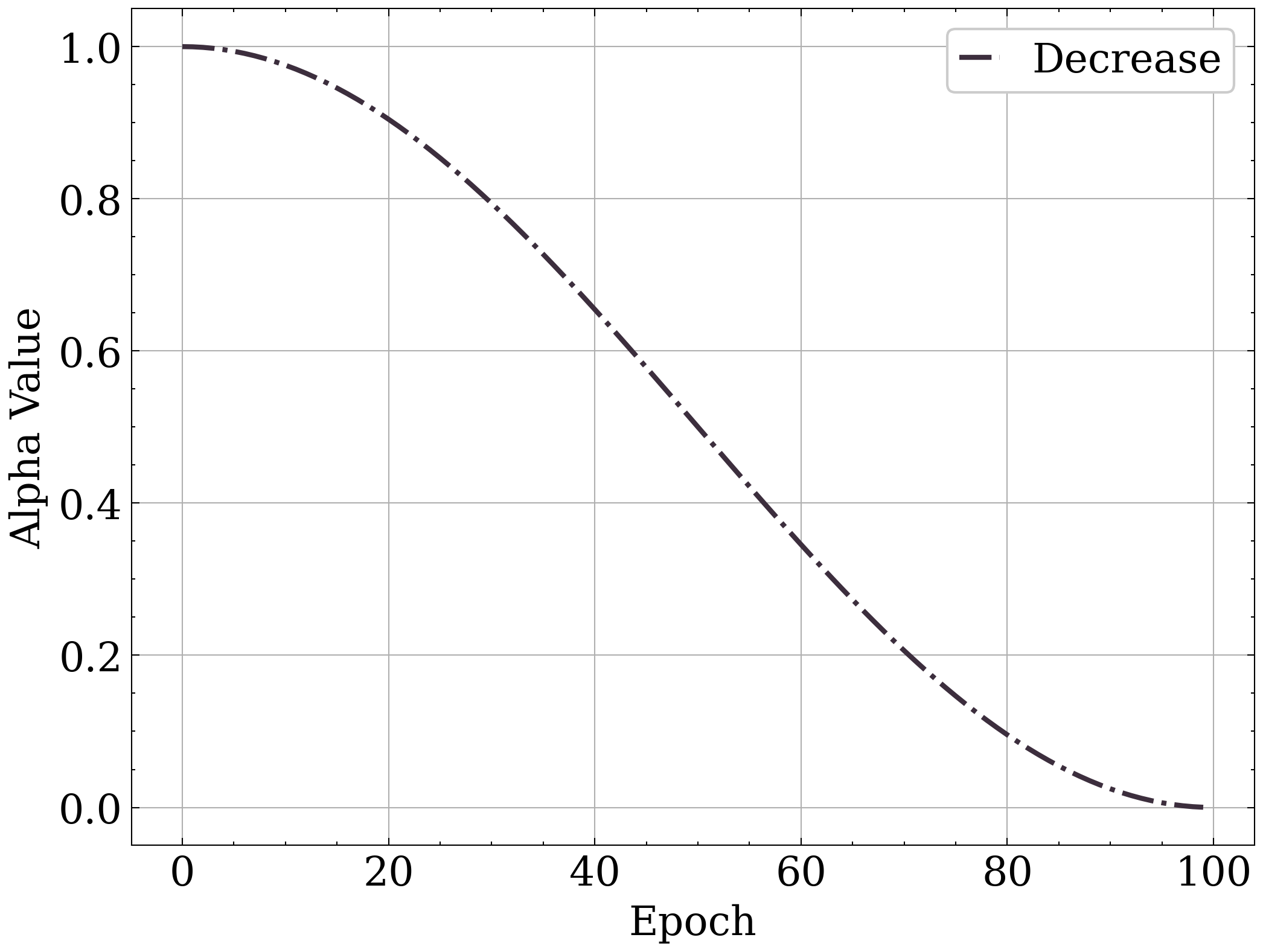}
		\label{fig:subfigure2}}
	\caption{The $\alpha$ value during the training process in four setting strategies, namely cosine annealing, cosine increment, cosine decrement and constant.}
	\label{fig:strategy}
\end{figure}

Here we introduce the $\alpha$ value setting details corresponding to cosine strategy and constant strategy adopted in this work as shown in Formula \ref{eq:fuse} and Formula \ref{eq:fuse_dec}.
The cosine annealing and cosine decrement strategies are calculated according to Formula \ref{alpha}.
To simplify the calculation, the cosine increment strategy is calculated according to Formula \ref{alpha_increase}.
In order to be more intuitive, we plotted the curve of $\alpha$ value during the training process, as shown in Figure \ref{fig:strategy}.

\begin{equation}\label{equ:cosine_incre}
	\alpha_{t+1} = 1 - \alpha_{min} - \frac{1}{2} (\alpha_{max} - \alpha_{min})(1-cos(\frac{T_{t}}{T_c}\pi))
	\label{alpha_increase}
\end{equation}

\subsection{Pre-trained Models Download}
All BERT pre-trainied models adopted in Table \ref{result_bert} are published by \citep{huggingface2018}.
In order to help reproduce our work and use our code easily, we summarize the download links of the pre-trained models as follows.

\para{PHOENIX-2014-T Dataset}
\begin{itemize}
  \item Multilingual: \emph{bert-base-multilingual-uncased}\\
  \url{https://huggingface.co/bert-base-multilingual-uncased}
  \item Distilbert: \emph{distilbert-base-german-cased} \\
  \url{https://huggingface.co/distilbert-base-german-cased}
  \item Gbert: \emph{gbert-base}\\
  \url{https://huggingface.co/deepset/gbert-base}
  \item Dbmdz: \emph{bert-base-german-dbmdz-uncased}\\
  \url{https://huggingface.co/bert-base-german-dbmdz-uncased}
\end{itemize}

\para{ASLG-PC12 Dataset}
\begin{itemize}
  \item Base-Tiny: \emph{bert-tiny}\\
  \url{https://huggingface.co/prajjwal1/bert-tiny}
  \item Electra: \emph{electra-small-discriminator}\\
  \url{https://huggingface.co/google/electra-small-discriminator}
  \item Distilbert: \emph{distilbert-base-uncased} \\
  \url{https://huggingface.co/distilbert-base-uncased}
  \item Base-uncased: \emph{bert-base-uncased} \\
  \url{https://huggingface.co/bert-base-uncased}
\end{itemize}


\end{document}